\documentclass[lettersize,journal]{IEEEtran}
\usepackage{amsmath,amsfonts}
\usepackage{algorithmic}
\usepackage{algorithm}
\usepackage{array}
\usepackage[caption=false,font=normalsize,labelfont=sf,textfont=sf]{subfig}
\usepackage{textcomp}
\usepackage{stfloats}
\usepackage{url}
\usepackage{verbatim}
\usepackage{booktabs}
\usepackage{makecell}
\usepackage{tabularx}
\usepackage{array} 
\usepackage{graphicx}
\usepackage[table]{xcolor}
\usepackage{amssymb}
 \usepackage{multirow}
\usepackage{subcaption}
\usepackage{cite}
\usepackage{caption}

\begin{document}

\title{Signal-SGN++: Topology-Enhanced Time–Frequency Spiking Graph Network for Skeleton-Based Action Recognition}

\author{Naichuan Zheng,Hailun Xia~\IEEEmembership{Member,~IEEE,},Weiyi Li,Yuchen Du
\thanks{Naichuan Zheng is with Beijing University of Posts and Telecommunications, School of Information and Communication Engineering, Beijing, China. Email: 2022110134zhengnaichuan@bupt.edu.cn.}     
\thanks{Hailun Xia is with Beijing University of Posts and Telecommunications, School of Information and Communication Engineering, Beijing, China. Email: hailunxia@bupt.edu.cn. Corresponding author.}
 \thanks{Weiyi Li is with Beijing University of Posts and Telecommunications, International School, Beijing, China. Email: liweiyi@bupt.edu.cn}
 \thanks{Yuchen Du is with Beijing University of Posts and Telecommunications, School of Information and Communication Engineering, Beijing, China. Email: duyuchen@bupt.edu.cn. Corresponding author.}
\thanks{This work is supported by The National Natural Science Foundation of China (NSFC) under grant No. 61976022.}}

\markboth{Under Review by IEEE Transactions on Neural Networks and Learning Systems}{}
\maketitle

\begin{abstract}
Graph Convolutional Networks (GCNs) demonstrate strong capability in modeling skeletal topology for action recognition, yet their dense floating-point computations incur high energy costs. Spiking Neural Networks (SNNs), characterized by event-driven and sparse activation, offer energy efficiency but remain limited in capturing the coupled temporal–frequency and topological dependencies of human motion. To bridge this gap, this article proposes Signal-SGN++, a topology-aware spiking graph framework that integrates structural adaptivity with time–frequency spiking dynamics. The network employs a backbone composed of 1D Spiking Graph Convolution (1D-SGC) and Frequency Spiking Convolution (FSC) for joint spatiotemporal and spectral feature extraction. Within this backbone, a Topology-Shift Self-Attention (TSSA) mechanism is embedded to adaptively route attention across learned skeletal topologies, enhancing graph-level sensitivity without increasing computational complexity. Moreover, an auxiliary Multi-Scale Wavelet Transform Fusion (MWTF) branch decomposes spiking features into multi-resolution temporal–frequency representations, wherein a Topology-Aware Time–Frequency Fusion (TATF) unit incorporates structural priors to preserve topology-consistent spectral fusion. Comprehensive experiments on large-scale benchmarks validate that Signal-SGN++ achieves superior accuracy–efficiency trade-offs, outperforming existing SNN-based methods and achieving competitive results against state-of-the-art GCNs under substantially reduced energy consumption.
\end{abstract}


\section{Introduction}
Human action recognition aims to classify human activities from motion patterns, and has broad applications in human--computer interaction, healthcare, and robotics~\cite{kong2022survey}. 
Among various data modalities, skeleton-based action recognition has attracted particular interest because it represents the human body as a set of joints (nodes) connected by bones (edges), providing a compact, viewpoint-robust description of motion\cite{sun2022review}. 
This graph-like representation allows models to focus on the underlying body structure rather than appearance variations such as clothing, background, or illumination.

Various architectures have been explored for skeleton-based action recognition, including Convolutional Neural Networks (CNNs)\cite{Du2015Skeleton,li2017skelcnn}, Recurrent Neural Networks(RNNs)\cite{du2015hrnn,liu2017gca,liu2017stlstm}, and Transformers\cite{xin2023transformer}. 
CNN-based methods are effective at local feature extraction but struggle to simultaneously capture long-range temporal dependencies and explicit joint-to-joint relationships. 
RNNs and LSTMs are naturally suited for sequential data, yet they often suffer from vanishing gradients and high computational cost on long sequences. 
Transformers excel at modeling long-range dependencies through self-attention, but when applied directly to skeleton sequences they typically ignore the inherent topological structure of the human body, thereby limiting their ability to model fine-grained joint interactions.

To explicitly encode this body structure, Graph Convolutional Networks (GCNs) have become a dominant paradigm for skeleton-based action recognition. 
In GCN-based methods, the human skeleton is modeled as a graph where joints are nodes and bones are edges, enabling direct reasoning over spatial dependencies and their temporal evolution\cite{feng2022gcnreview}. 
ST-GCN was the first to introduce a spatial--temporal GCN framework for skeleton-based action recognition, jointly modeling joint configurations and motion patterns\cite{Yan2018Spatial}. 
Subsequent works, such as 2S-AGCN\cite{Shi2019Two} and CTR-GCN\cite{Chen2021Channel}, further enhanced performance by designing adaptive graph constructions and attention mechanisms to capture dynamic joint dependencies, while Block-GCN improved efficiency by partitioning the skeleton into subgraphs for more focused modeling of local spatial patterns\cite{Zhou2024BlockGCN}. 
These advances highlight the importance of leveraging the intrinsic skeletal topology for robust and accurate action recognition.

Despite their effectiveness, GCN-based models are computationally expensive due to their reliance on dense matrix operations and floating-point computations (FLOPs), resulting in high energy consumption, which makes them less suitable for real-time applications, particularly in resource-constrained environments such as mobile and edge devices. In contrast, Spiking Neural Networks (SNNs), as third-generation neural networks, utilize binary spike-based encoding to emulate neural dynamics. Unlike traditional artificial neural networks (ANNs) that process information continuously, SNNs operate in an event-driven manner, where neurons fire only in response to events (spikes)\cite{maass1997spiking}. This sparse activation leads to significantly lower energy consumption and greater computational efficiency\cite{roy2019spike,schuman2022neuromorphic}.

In recent years, several SNN-based models have been introduced for various visual tasks\cite{singhal2025snnreview}. The first significant breakthrough is Spikformer\cite{Zhou2022Spikformer}, which replaces traditional floating-point operations with binary spike-based self-attention, drastically reducing energy consumption. This led to further advancements with multiple variants of spiking Transformers optimizing energy efficiency\cite{Yao2023SpikeTransformer,Yao2024SpikeV2,Fang2024SWformer,Lee2025STAtten}. Meanwhile, SpikingGCN combined SNNs with graph convolutions, using a Bernoulli pulse encoder to convert continuous features into discrete spikes, enabling the first integration of SNNs with graph convolutions\cite{Zhu2022SpikingGCN}. Following this, Dy-SIGN proposed a dynamic spiking graph framework that integrates information compensation and implicit differentiation to enhance efficiency\cite{Yin2024DySIGN}. More recently, MK-SGN combined multimodal fusion, self-attention, and knowledge distillation, creating an energy-efficient Spiking GCN framework for skeleton-based action recognition\cite{Zheng2025MKSGN}.

Despite these advancements, current SNN models still face significant limitations in fully capturing the dynamic topology of skeleton data. These models typically treat the entire skeleton sequence as a static image, adding a pulse step dimension, but fail to adapt to the dynamic inter-joint relationships during an action. Furthermore, while SNNs mimic neural spike activity and leverage spiking-based attention, these attention mechanisms lack the flexibility to effectively learn and adapt to the evolving joint dependencies and topological structures. Additionally, although pulse signals inherently exhibit non-stationarity and time-frequency characteristics, similar to EEG signals, current SNN models often overlook these properties\cite{AlFahoum2014EEG, Morales2022TimeFrequency}. Consequently, this oversight results in inadequate modeling of the dynamic changes in joint activity over time, especially in capturing frequency-domain features. Consequently, existing SNN models struggle with time-frequency feature extraction and fusion, limiting their ability to capture the fine-grained dynamic variations in human actions.
\begin{figure}[t]
    \centering
    \includegraphics[width=0.5\textwidth]{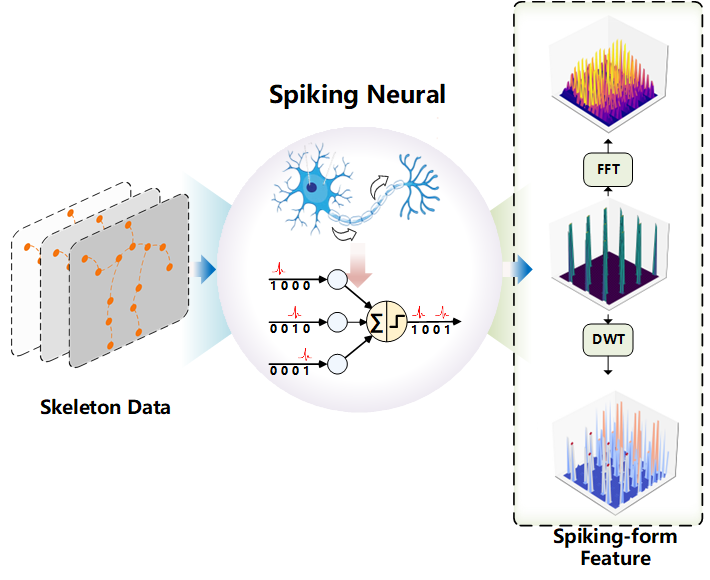} 
    \caption{Motivating Perspective of Signal-SGN++. Skeleton data is encoded through a spiking neural network, transforming it into a sparse binary representation that is interpreted as a multi-dimensional discrete signal. This approach enables the application of time-frequency analysis, focusing on the extraction of spectral-temporal patterns, which is a critical component for enhancing action recognition performance in Signal-SGN++.}
    \label{fig:sgn} 
\end{figure}
To address the challenges faced by existing methods in skeleton-based action recognition, this article proposes Signal-SGN++, an enhanced topology-aware spiking graph network designed to capture both temporal–frequency dynamics and topological dependencies efficiently. A recent conference work Signal-SGN ~\cite{2025signalsgn} explored time–frequency spiking graph modeling for this task. Building on that line of research, Signal-SGN++ introduces several key designs to improve both recognition performance and energy efficiency further. Specifically, the framework consists of three primary modules: 1D Spiking Graph Convolution (1D-SGC), Topology-Shift Self-Attention (TSSA), and Frequency Spiking Convolution (FSC). The 1D-SGC module applies spiking graph convolutions to skeleton sequences, enabling efficient extraction of spatial and temporal features while preserving the energy efficiency inherent to SNNs. TSSA, integrated within 1D-SGC, introduces an adaptive attention mechanism guided by the learned skeletal adjacency, allowing the model to focus on informative joints and capture evolving inter-joint relationships during dynamic actions. The FSC module leverages Fast Fourier Transform (FFT) to transform spiking signals into the frequency domain, facilitating the extraction of frequency-specific features crucial for recognizing complex actions with subtle temporal variations.

In addition to these components, Signal-SGN++ incorporates a Multi-Scale Wavelet Transform Feature Fusion (MWTF) branch with a Topology-Aware Time–Frequency Fusion (TATF) unit. MWTF decomposes features into multi-resolution frequency components to capture both high- and low-frequency dynamics, while TATF fuses these components in a topology-consistent manner to enhance fine-grained temporal–frequency modeling. By combining these modules, Signal-SGN++ dynamically adapts to changing inter-joint relationships and provides an efficient and accurate solution for skeleton-based action recognition.

The major contributions of this article are summarized as follows:
\begin{enumerate}
\item We propose Signal-SGN++, a novel topology-aware spiking graph network that integrates 1D-SGC, TSSA, and FSC to effectively capture temporal-frequency dynamics and topological dependencies in skeleton-based action recognition.

\item We introduce the MWTF module, which, in conjunction with TATF, enhances the model’s ability to capture both high- and low-frequency dynamics in a topology-consistent manner.

\item Extensive experiments on benchmark datasets demonstrate that Signal-SGN++ outperforms existing methods in terms of both accuracy and energy efficiency, offering a more effective solution for skeleton-based action recognition compared to state-of-the-art GCN and SNN models.
\end{enumerate}
This article builds upon a prior anonymous conference submission on Signal-SGN~\cite{2025signalsgn}. The additional contributions are summarized as follows:
\begin{enumerate}
\item Topology-Shift Self-Attention (TSSA) is introduced to improve the model’s ability to capture dynamic joint dependencies and topological relationships, which significantly enhances the model's performance in skeleton-based action recognition.

\item Topology-Aware Time-Frequency Fusion (TATF) is proposed and integrated into the Multi-Scale Wavelet Transform Feature Fusion (MWTF) module, ensuring topology-consistent fusion of temporal-frequency components, which further improves the model's ability to capture fine-grained action dynamics.

\item Extensive experiments comparing TSSA with other spiking-based attention mechanisms, including the original SSA, demonstrate that TSSA outperforms the alternatives, providing stronger representation of joint dependencies.

\item Theoretical enhancements are made to deepen the understanding of the temporal, spatial, and frequency-specific dynamics in the model, providing a more robust framework for skeleton-based action recognition.

\item A more comprehensive visualization analysis is conducted, including visualizing the learned topological structures and the model's feature response maps. This analysis provides a detailed understanding of the mechanisms behind Signal-SGN++, highlighting the current challenges in dynamic skeleton modeling and suggesting potential areas for future research.
\end{enumerate}

\section{Related Works}
\subsection{Skeleton-based Action Recognition}
Skeleton-based action recognition has evolved from early CNN-based methods that treat skeletons as pseudo-images\cite{Du2015Skeleton} to more sophisticated architectures. Attention mechanisms has been introduced to focus on discriminative joints \cite{Si2019Attention} and Transformer models has been used to capture long-range dependencies \cite{Plizzari2021Spatial}. A significant paradigm shift occurred with the introduction of Graph Convolutional Networks (GCNs), pioneered by ST-GCN\cite{Yan2018Spatial}, which directly operate on the non-Euclidean skeleton structure. Subsequent advancements have focused on refining GCN architectures, including learning data-driven adaptive graphs (2s-AGCN\cite{Shi2019Two}), improving computational efficiency with shift operations (Shift-GCN \cite{Cheng2020Skeleton} ), integrating multi-scale convolutions (MS-G3D \cite{Liu2020Disentangling}), learning channel-wise topologies (CTR-GCN \cite{Chen2021Channel}), embeding skeleton information into the latent representations of human action.  (InfoGCN\cite{Li2021InfoGCN} ), and extracting more semantic connections through hierarchical decomposition (HD-GCN \cite{Lee2023Hierarchically}). Recent works like BlockGCN \cite{Zhou2024BlockGCN}  have further refined topology awareness and DEGCN \cite{myung2024degn} introduces deformable temporal convolution for skeleton-based action recognition. Large-foundation models like Hulk \cite{Wang2025Hulk} unify skeleton recognition with other human-centric tasks. Despite this progress, these Artificial Neural Network (ANN) based models are computationally intensive, limiting their deployment on energy-constrained edge devices.
\subsection{Spiking Neural Networks}
Spiking Neural Networks (SNNs) offer an energy-efficient alternative to ANNs by processing information with sparse, discrete spike trains, making them suitable for neuromorphic hardware. The dynamics of a typical SNN neuron are often described by the Leaky Integrate-and-Fire (LIF) model \cite{Gerstner2014Neuronal}, where the membrane potential $u(t)$ evolves according to the differential equation:
\begin{equation}
    \tau_m \frac{du(t)}{dt} = -[u(t) - u_{\text{rest}}] + R \cdot I(t),
\end{equation}
and when $u(t)$ reaches a threshold $\theta$, the neuron fires a spike and its potential is reset.
Recent research has focused on adapting the powerful Transformer architecture for SNNs. Spikformer\cite{Yao2023SpikeTransformer} introduced a Spiking Self-Attention (SSA) mechanism, while the Spike-driven Transformer series\cite{Yao2023SpikeTransformer,Yao2024SpikeV2} developed a multiplication-free self-attention operator for extreme hardware efficiency. Other works have explored saccadic attention  \cite{Zou2025SNNViT}, block-wise spatial-temporal attention mechanism\cite{Lee2025STAtten}, and Q-K-only linear-complexity attention\cite{Zhou2024QKFormer} to maintain spike sparsity and efficiency. And Fang et al. \cite{Fang2024SWformer} propose the Spiking Wavelet Transformer (SWformer), which uses a Frequency-Aware Token Mixer (FATM) to learn spatial-frequency features without any attention mechanism. For graph-structured data, SpikingGCN \cite{Zhu2022SpikingGCN} pioneered the integration of GCNs with SNNs, and subsequent works like Dy-SIGN\cite{Yin2024DySIGN} and MK-SGN \cite{Zheng2025MKSGN} have introduced dynamic graph processing and multimodal fusion, achieving significant energy savings. However, a key limitation of existing SNN-based models is their inadequate modeling of dynamic skeletal topologies and their failure to fully leverage the rich time-frequency characteristics inherent in spike-based representations\cite{AlFahoum2014EEG, Morales2022TimeFrequency}, which are analogous to biological neural signals.
\section{Method}
\begin{figure*}[t]
    \centering
\includegraphics[width=1\textwidth]{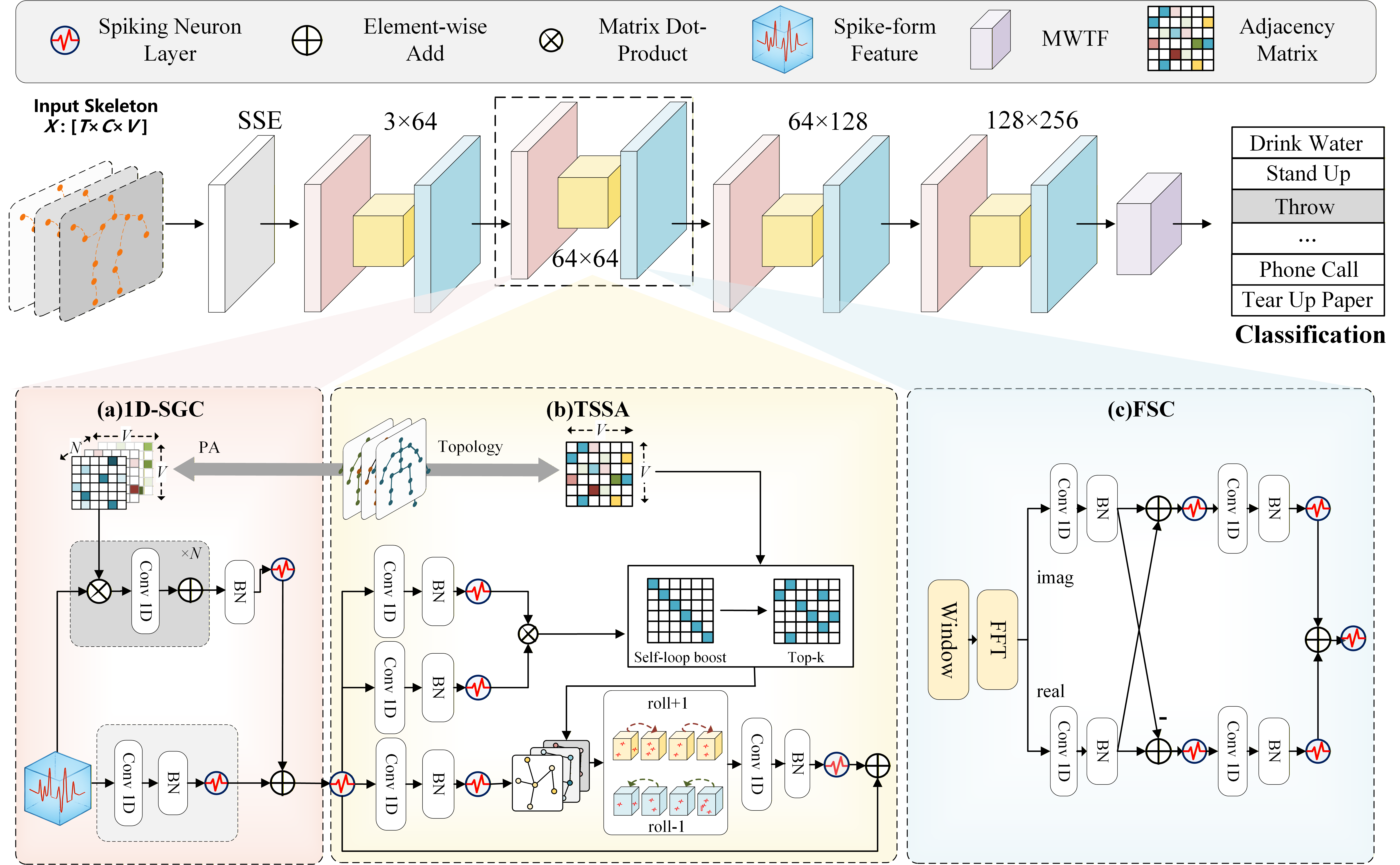}
    \caption{Signal-SGN++ Architecture. The model consists of a series of stacked modules: 1D-Spiking Graph Convolution (1D-SGC), followed by Topology-Shift Self-Attention (TSSA), then Frequency Spiking Convolution (FSC), and finally the Multi-Scale Wavelet Transform Fusion (MWTF) for feature fusion and classification.}
    \label{fig:Signal-SGN++}
\end{figure*}
The proposed Signal-SGN++ framework, as illustrated in Figure~\ref{fig:Signal-SGN++}, is designed to capture both spatiotemporal and frequency-domain features for skeleton-based action recognition. First, Skeleton Spike Encoding (SSE) is applied to convert the input skeleton sequence into spiking signals, ensuring consistent encoding of joint movements. The 1D-SGC module, combined with Topology-Shift Self-Attention (TSSA), is then used to model spatial dependencies and dynamically adjust joint relationships. Following this, the FSC module leverages FFT to transform spiking features into the frequency domain, enabling frequency-aware learning. An auxiliary MWTF branch decomposes the features into multi-resolution temporal-frequency representations, which are fused using the TATF unit to maintain topology-consistent spectral fusion. Finally, the fused features are processed through Global Average Pooling (GAP) and a fully connected (FC) classification head to produce the final prediction. The details of each component are explained in the subsequent sections.

\subsection{Skeleton Spike Encoding (SSE)}
\label{subsec:SSE}
The input skeleton sequence \(X \in \mathbb{R}^{C \times T \times V}\) contains 3D joint coordinates that may lie in different spatial quadrants and thus include both positive and negative values. Since SNNs are typically designed for non-negative inputs and binarization can easily discard information, we first perform a simple per-channel min–max normalization on each coordinate dimension \((x,y,z)\) over the entire sequence, linearly mapping all values into \([0,1]\) while preserving their relative geometric structure. The normalized skeleton input is then processed through a 1D Convolution (Conv1D), followed by Batch Normalization (BN) and a Spiking Neuron (SN) layer, forming the initial encoding stage of Signal-SGN++. The resulting feature representation is denoted as \( X_{\text{i}} \). 
\subsection{1D-SGC Module}
\label{subsec:1D-SGC}

In the 1D-SGC module, inspired by Graph Convolution Networks (GCNs), topological information is integrated by normalizing adjacency-based weights to aggregate features from neighboring joints, as shown in Figure~\ref{fig:Signal-SGN++}(a). The skeleton is represented as a graph \( \mathcal{G}(V, E) \), where joints form the vertices \( V \) and bones form the edges \( E \). This structure is captured in the adjacency matrix \( A \in \mathbb{R}^{V \times V} \), where \( A_{u,v} = 1 \) if joints \( u \) and \( v \) are physically connected, and \( A_{u,v} = 0 \) otherwise.

To incorporate topological relationships into the feature aggregation, the adjacency matrix is normalized as $\mathbf{A}_{\text{norm}}= \Delta^{-\frac{1}{2}} \widetilde{A} \Delta^{-\frac{1}{2}}$,
where \( \widetilde{A} = A + I \) adds self-connections \( \), and \( \Delta \) is the degree matrix of \( \widetilde{A} \). This normalization ensures that the influence of each node is appropriately scaled according to its connectivity, facilitating effective integration of topological information.

To enable dynamic topology learning, we introduce a set of learnable adjacency matrices $\text{PA} \in \mathbb{R}^{N \times V \times V}$, which are initialized using the powers of the normalized adjacency matrix: $\text{PA}[n, :, :] \leftarrow \mathbf{A}_{\text{norm}}^n$. Then, Conv1D and BN are applied along the channel dimension to extract features for each skeleton joint at each time step. This allows for feature aggregation across the skeleton’s joints, as represented by the following equation:
\begin{equation} G_\text{i} = \mathrm{BN}\left[\sum_{n=0}^{N} \operatorname{Conv1d}\left(X_\text{i} * \text{PA}[n, :, :]\right)\right], \end{equation}
where $\text{PA}[n, :, :]$ denotes the $n$-th learnable relational matrix, $N$ denotes the number of relational matrices in the decomposition, and $*$ denotes matrix multiplication. The output $G_\text{i} \in \mathbb{R}^{T \times D \times V}$, where $D$ is the hidden channel dimension, captures the aggregated features.

To leverage the temporal dynamics in Spiking SNNs, the temporal dimension of the features is mapped to the spike time steps, and the features are input into SN. The updated feature is then computed as:

\begin{equation}
G^{(0)}_\text{o} = SN\left(G_\text{i}\right) + SN\left(\mathrm{BN}(\operatorname{Conv1d}(X_\text{i}))\right).
\end{equation}
\subsection{TSSA Module}
While the 1D-SGC module effectively integrates topological information through normalized adjacency-based aggregation, it processes all joints uniformly without considering their dynamic importance during different actions. Building upon the spiking self-attention mechanism proposed by Zhou et al. \cite{Zhou2022Spikformer} in Spikformer, we introduce the TSSA module, as shown  in Figure \ref{fig:Signal-SGN++}(b), to address two critical challenges: the quadratic computational complexity with respect to the number of joints ($O(V^2)$), and the neglect of physical topology encoded in the adjacency matrix $S$. TSSA introduces a fused neighbor selection strategy combining topological structure with dynamic feature similarity, and replaces matrix multiplication with lightweight channel shift operations.

We first apply linear projections to the input features $G^{(0)}_\text{o}$ from the 1D-SGC module to generate Query, Key, and Value representations through Conv1d, BN, and SN. The projected features are then reshaped into multi-head form with $H$ attention heads, each having dimension $d^{(0)} = D^{(0)}/H$. After obtaining the Q and K matrices, we compute the feature similarity between joints via dot product for each time step $t$ and attention head $h$, followed by averaging across all heads to obtain the similarity matrix $\mathcal{Q}_t^{(0)} \in \mathbb{R}^{V \times V}$.

To leverage the dynamic topological structure learned in the 1D-SGC module, we introduce an adaptive topology adjacency score matrix $\mathcal{A}$. This matrix is derived from the learnable adaptive adjacency matrices $\text{PA} \in \mathbb{R}^{N \times V \times V}$ (as defined in Section \ref{subsec:1D-SGC}), which are optimized during training to capture action-specific joint correlations. The topology score matrix $\mathcal{A}$ is computed by aggregating the correlation strengths across all relation types:
\begin{equation}
\label{equation:PA}
\mathcal{A} = \sum_{n=0}^{N-1} |\text{PA}[n, :, :]| \in \mathbb{R}^{V \times V}
\end{equation}
where  $|\cdot|$ denotes element-wise absolute value. By aggregating correlation strengths across all relation types, $\mathcal{A}$ encodes the learned topological importance of joint connections.

Subsequently, we perform weighted fusion of the adaptive topology score $\mathcal{A}$ derived from $\text{PA}$ and the feature similarity matrix $\mathcal{Q}_t^{(0)}$:
\begin{equation}
\mathcal{C}_t^{(0)} = \alpha \mathcal{A} + (1-\alpha) \mathcal{Q}_t^{(0)} \in \mathbb{R}^{V \times V}
\end{equation}
$\alpha$ is a learnable parameter that balances the static topological prior and dynamic feature relationships, and is initialized to 0.7 in our experiments. After obtaining the fused score matrix, we select the Top-$k$ source joints with the highest scores for each target joint $j$ to construct a sparse neighbor set $\mathcal{N}_{j,t}^{(0)}$, where the neighbors are ranked by their fusion scores in descending order. To ensure the inclusion of self-loops, we explicitly assign the highest score to each joint's self-connection.

In the feature aggregation stage, we introduce the channel shift operation. We define $\sigma_+$ as rightward cyclic shift ($\text{roll+1}$) and $\sigma_-$ as leftward cyclic shift ($\text{roll-1}$), which are applied alternately to the feature vectors of the $k$ selected neighbors. For each target joint $j$, attention head $h$, and time step $t$, we aggregate the shifted features:

\begin{equation}
g_{j,h,t}^{(0)} = \sum_{r=0}^{k-1} \text{Shift}\left(v_{\mathcal{N}_{j,t}^{(0)}[r], h,t}^{(0)}, \sigma_r\right) \in \mathbb{R}^{d^{(0)}}
\end{equation}
where $v_{i,h,t}^{(0)} \in \mathbb{R}^{d^{(0)}}$ denotes the Value feature vector of source joint $i$ from the V projection, $\mathcal{N}_{j,t}^{(0)}[r]$ denotes the $r$-th neighbor of joint $j$, and $\sigma_r$ alternates between $\sigma_+$ and $\sigma_-$ based on the parity of $r$. The outputs from all heads are concatenated to obtain the multi-head aggregated features $g_{j,t}^{(0)} \in \mathbb{R}^{D^{(0)}}$, where $D^{(0)} = H \cdot d^{(0)}$. Finally, we apply an output projection through Conv1d, BN, and SN to the aggregated features, followed by a residual connection with the original input $G_{\text{o},t}^{(0)}$ to facilitate gradient flow. The complete temporal sequence output is denoted as $G^{(0)} \in \mathbb{R}^{T \times V \times D^{(0)}}$.

Through the above design, the TSSA module achieves three key advantages: (1) leveraging both physical priors and learned action-specific patterns by fusing the adaptive topological structure encoded in $\text{PA}$ with dynamic feature similarity, (2) reducing computational complexity from $O(V^2 \cdot d^{(0)})$ to $O(k \cdot V \cdot d^{(0)})$ through Top-$k$ sparsification, and (3) enhancing numerical stability and energy efficiency by replacing matrix multiplication with channel shifts.

\subsection{FSC Module}

Following the TSSA module, which captures both topological structure and dynamic feature relationships in the spatial domain, we introduce the Frequency-domain Spiking Convolution (FSC) module to further extract meaningful frequency-domain features from the attention-enhanced representations. The TSSA output $G^{(0)} \in \mathbb{R}^{T \times V \times D^{(0)}}$ consists of spiking-form features—discrete, aperiodic signal sequences encoded as binary values (0s and 1s). As shown in Figure \ref{fig:Signal-SGN++}(c), the FSC module transforms these spatial-domain features into the frequency domain to capture complementary spectral patterns across joints.

First, this module applies a learnable window function \(w_V(v) \in \mathbb{R}^{V}\) along the joint dimension to mitigate spectral leakage and enhance frequency analysis. Unlike fixed windows, \(w_V\) assigns adaptive weights to each joint, acting as a soft spatial gate before the frequency transform and producing the windowed feature
\(F^{(0)}_{\text{i}} = G^{(0)} \cdot w_V(v)\),
where \(\cdot\) denotes broadcasting over the joint dimension and \(F^{(0)}_{\text{i}}\) is fed into the frequency encoder. An FFT is then performed along \(V\) to project \(F^{(0)}_{\text{i}}\) into the frequency domain; the resulting complex-valued representation is split into real and imaginary parts, \(F^{(0)}_{\text{R}}\) and \(F^{(0)}_{\text{I}}\), which are processed by parallel branches. This joint-wise frequency transform enables the model to capture interpretable spatial–frequency patterns across joints, leading to enhanced action representation.

To process the complex-valued FFT outputs, we propose a spiking complex convolution module that preserves and utilizes amplitude and phase information, enhancing the network's ability to capture detailed frequency-domain features:
\begin{equation}
\begin{aligned}
F^{(0)}_{\text{o},\text{R}}= \mathrm{BN}[\operatorname{Conv1d}(F^{(0)}_{\text{R}})],
F^{(0)}_{\text{o},\text{I}}= \mathrm{BN}[\operatorname{Conv1d}(F^{(0)}_{\text{I}})],
\end{aligned}
\end{equation}
\begin{equation}
F^{(0)}_{\text{o},u} = \text{SN}\left(F^{(0)}_{\text{o},\text{R}} \pm_u F^{(0)}_{\text{o},\text{I}}\right), \quad u \in \{1, 2\}.
\end{equation}
where $\pm_u$ represents addition ($+$) when $u = 1$, and subtraction ($-$) when $u = 2$. The fused output is then computed as:
\begin{equation}
F^{(0)} = \text{SN}(\text{BN}(\text{Conv1d}(F^{(0)}_{\text{o},1}))) + \text{SN}(\text{BN}(\text{Conv1d}(F^{(0)}_{\text{o},2}))).
\end{equation}

The output $F^{(0)}$ represents the extracted frequency-domain features, serving as an essential complement to the spatial-domain features from the TSSA module. A complete 1D-SGC-TSSA-FSC process can then be expressed as:
\begin{equation}
X^{(l)} = F^{(l)} + G^{(l)} + \operatorname{res}\left(X^{(l-1)}\right)
\end{equation}
where $X^{(l)}$ denotes the output of the $l$-th Signal-SGN++ block, combining the frequency-domain features $F^{(l)}$ from the FSC module, the spatial-domain features $G^{(l)}$ from the TSSA module, and a residual connection from the previous layer $\operatorname{res}\left(X^{(l-1)}\right)$. This design enables the network to leverage both spatial topology-aware attention and frequency-domain representations for comprehensive action recognition.

\subsection{Multi-Scale Wavelet Transform Feature Fusion Module}

\begin{figure*}[!htbp]
\centering
\includegraphics[width=1\textwidth]{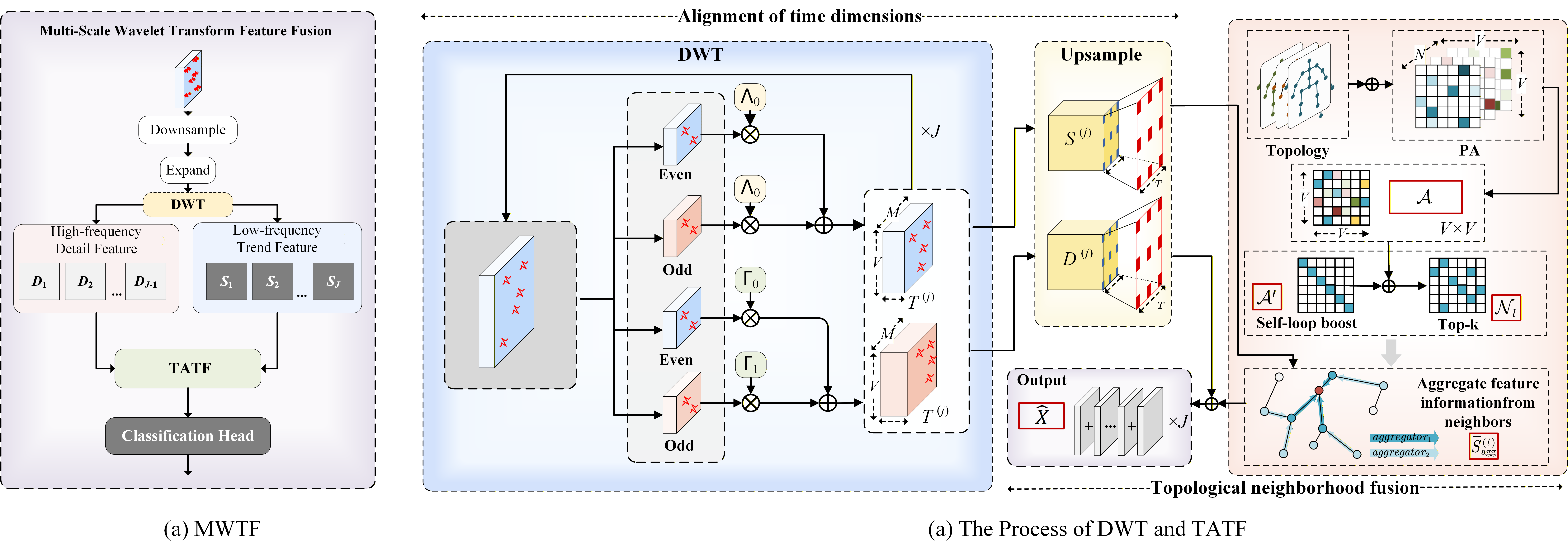}
\caption{Multi-Scale Wavelet Transform and Topological Fusion.
(a) The overall framework of the MWTF module, which integrates multi-scale wavelet transforms and topological fusion.
(b) The complete process from DWT to TATF, including the alignment of time dimensions, decomposition into high-frequency and low-frequency components, and topological neighborhood fusion.}
\label{fig:MWTF}
\end{figure*}

We formulate the MWTF module to capture the temporal dynamics and frequency-specific characteristics of spiking-form features, with its outputs fused with backbone outputs to enhance classification performance, as illustrated in Figure \ref{fig:MWTF}. Traditional wavelet transform methods typically process detail and approximation components through simple addition or independent processing, ignoring spatial topological structure. In contrast, our MWTF module introduces a selective topology-aware fusion strategy that leverages the physical topology of human skeletons to enhance time-frequency features.

\subsubsection{Legendre Polynomial-Based Filter Construction}

We construct wavelet filters based on Legendre polynomials \(P_k(x)\), which provide a robust orthogonal basis for decomposing spike-like features in temporal signals. The polynomials are generated recursively using the relation:
\begin{equation}
\begin{aligned}
& P_0(x) = 1, \quad P_1(x) = x, \\
(m+1) P_{m+1}(x) &= (2m+1)xP_m(x) - mP_{m-1}(x).
\end{aligned}
\end{equation}
Their orthogonality condition ensures that different polynomial orders yield mutually independent basis functions. To apply these polynomials to temporal-domain signals sampled over \([0,T]\), we first normalize the time index \(u = t/T \in [0,1]\), then map it to \([-1,1]\) via \(x = 2u - 1\). This transformation aligns the original time domain with the canonical domain of the Legendre polynomials.

Low-frequency (low-pass) filters are derived from lower-order Legendre polynomials, which vary slowly across \([-1,1]\). Each polynomial is scaled by \(\sqrt{2m+1}\) for proper normalization. Introducing a half-period shift \(\Delta = 0.5\) in the normalized time domain allows us to construct two distinct low-pass filters:
\begin{equation}
\begin{aligned}
\Lambda_0[m,t] = \sqrt{2m+1}\,P_m(2u-1), \\
\quad \Lambda_1[m,t] = \sqrt{2m+1}\,P_m(2u).
\end{aligned}
\end{equation}

For high-frequency (high-pass) filters, we derive them from the low-pass filters with appropriate scaling:
\begin{equation}
\begin{aligned}
\Gamma_0[m,t] = \sqrt{2}\,\Lambda_0[m,t],\\
\quad \Gamma_1[m,t] = \sqrt{2}\,\Lambda_1[m,t].
\end{aligned}
\end{equation}

These four filters \((\Lambda_0,\Lambda_1,\Gamma_0,\Gamma_1)\) decompose the signal into distinct frequency bands, enabling a nuanced analysis of spike-like features. The parameter \(M\) specifies the number of frequency coefficients extracted, influencing the spectral resolution. A more detailed derivation of these formulations is provided in the appendix.

\subsubsection{Channel Downsampling and Temporal Padding}

Given the input features \(X^{(L)} \in \mathbb{R}^{T \times C^{(L)} \times V}\) from the final layer \(L\) of the backbone network, we apply spiking neuron activation followed by two-stage channel downsampling. First, we apply grouped convolution to reduce the channel dimension from \(C^{(L)}\) to an intermediate dimension \(C_{\text{mid}}\), and subsequently apply standard 2D convolution and batch normalization to further reduce the channel dimension to \(M\), yielding the downsampled feature \(X^{(L)}_{\text{down}} \in \mathbb{R}^{T \times M \times V}\), where \(M \ll C^{(L)}\) ensures computational efficiency and aligns with the number of frequency coefficients in the wavelet filters. To ensure compatibility with the wavelet transform, we pad the temporal dimension \(T\) to the nearest power of two: \(T_{\text{padded}} = 2^{\lceil \log_2 T \rceil}\). The extended input \(\widetilde{X} = [X^{(L)}_{\text{down}}, X_{\text{extra}}]\) is formed by concatenating duplicated copies of the beginning frames along the temporal axis.

\subsubsection{Multi-Wavelet Decomposition}

In this section, we describe the multi-wavelet decomposition of the input $\widetilde{X}$ using wavelet filters $\Lambda_0$, $\Lambda_1$, $\Gamma_0$, and $\Gamma_1$, which produces detail coefficient $D^{(j)}$ for high-frequency features and scaling coefficients $S^{(j)}$ for low-frequency trends, as shown in the left of Figure\ref{fig:MWTF} (b). 

The extended input \(\widetilde{X}\) is then recursively decomposed using the constructed wavelet filters \(\Lambda_0\), \(\Lambda_1\), \(\Gamma_0\), and \(\Gamma_1\), yielding two types of outputs at each level: detail coefficients \(D^{(j)}\), which capture high-frequency, localized temporal features, and scaling coefficients \(S^{(j)}\), which preserve the lower-frequency trend components. At each decomposition level \(j\), we perform even-odd sampling to split the time series into even and odd frames along the temporal axis. Specifically, the even-indexed frames \(X_{\text{even}}^{(j)} = \widetilde{X}^{(j)}[::2, :, :]\) and odd-indexed frames \(X_{\text{odd}}^{(j)} = \widetilde{X}^{(j)}[1::2, :, :]\) are extracted. The detail and scaling coefficients are then computed by applying the high-pass and low-pass filters:

\begin{equation}
\begin{aligned}
D^{(j)} &= \Gamma_0 \cdot X_{\text{even}}^{(j)} + \Gamma_1 \cdot X_{\text{odd}}^{(j)} \in \mathbb{R}^{T^{(j)} \times M \times V} \\
S^{(j)} &= \Lambda_0 \cdot X_{\text{even}}^{(j)} + \Lambda_1 \cdot X_{\text{odd}}^{(j)} \in \mathbb{R}^{T^{(j)} \times M \times V}
\end{aligned}
\end{equation}
where \(T^{(j)} = T^{(j-1)}/2\) represents the temporal length at level \(j\), and \(\cdot\) denotes matrix multiplication applied along the channel dimension. The scaling coefficient \(S^{(j)}\) is used as the input for the next decomposition level by setting \(\widetilde{X}^{(j+1)} = S^{(j)}\), while the detail coefficient \(D^{(j)}\) is retained for subsequent fusion. 

\subsubsection{Topology-Aware Time-Frequency Fusion}
To enable multi-scale fusion, all detail and scaling coefficients are first upsampled to a unified temporal length \(T\) via linear interpolation, yielding \(D^{(j)}_{\text{up}}, S^{(j)}_{\text{up}} \in \mathbb{R}^{T \times M \times V}\). As shown on the right side of Fig.~\ref{fig:MWTF}(b), these upsampled coefficients are then fed into the proposed TATF module, which exploits the action-specific topology learned by the final backbone layer to capture dependencies between high- and low-frequency components. In particular, TATF selectively enhances low-frequency scaling coefficients along topological neighborhoods while preserving high-frequency detail coefficients. Compared with conventional cross-attention with quadratic complexity \(O(V^2)\), this topology-guided aggregation operates with linear complexity \(O(k_{\text{topo}} \cdot V)\), where \(k_{\text{topo}} \ll V\).

The topology score matrix \(\mathcal{A} \in \mathbb{R}^{V \times V}\) is derived from the TSSA module in the last backbone layer. As defined in Eq.~\eqref{equation:PA}, it is obtained by aggregating correlation strengths across all relation types of the adaptive adjacency matrices \(\text{PA} \in \mathbb{R}^{N \times V \times V}\), where \(\mathcal{A}[u,v]\) encodes the topological correlation from source joint \(u\) to target joint \(v\). To ensure that each joint prioritizes its self-connection, we enhance the diagonal entries and form an augmented topology matrix \(\mathcal{A}'\):
\begin{equation}
\mathcal{A}'[v,v] = \max_{u \in \{1,\ldots,V\}} \mathcal{A}[u,v] + 1.0.
\end{equation}
For each target joint \(v\), we then select the Top-\(k_{\text{topo}}\) source joints with the highest augmented scores \(\mathcal{A}'[u,v]\) to construct the neighbor set \(\mathcal{N}_v\), where \(k_{\text{topo}} = 6\) in our implementation. This guarantees that \(\mathcal{N}_v\) always contains \(v\) itself (due to the diagonal enhancement), together with the \(k_{\text{topo}}-1\) most relevant neighbors.
Finally, topology-aware aggregation is applied to the low-frequency scaling coefficients through mean-normalized neighborhood averaging:
\begin{equation}
\bar{S}^{(j)}_{\text{agg}}[t, c, v] = \frac{1}{k_{\text{topo}}} \sum_{u \in \mathcal{N}_v} S^{(j)}_{\text{up}}[t, c, u],
\end{equation}
where \(t\) denotes the time step, \(c\) the channel index, and \(v\) the target joint. This operation propagates structural information along the learned skeleton topology while keeping the original high-frequency details intact, and the mean normalization helps maintain feature scale consistency and numerical stability during training.
With the aggregated low-frequency features $\bar{S}^{(j)}_{\text{agg}}$, we can now formulate the fusion of
 detail and scaling coefficients at each decomposition level:
\begin{equation}
X^{(j)}_{\text{fused}} = D^{(j)}_{\text{up}} + \lambda \cdot \bar{S}^{(j)}_{\text{agg}} \in \mathbb{R}^{T\times M \times V}
\end{equation}
where \(\lambda\) initialized to 0.1, is a learnable strength coefficient, and \(\bar{S}^{(j)}_{\text{agg}}\) is the mean-normalized neighborhood aggregation result of the low-frequency scaling coefficients.
Compared with the conventional cross-attention used in Signal-SGN~\cite{2025signalsgn}, TATF is both more efficient and better aligned with the learned topology. By restricting aggregation to Top-\(k_{\text{topo}}\) neighbors, it reduces the complexity from \(O(V^2 \cdot d)\) to \(O(k_{\text{topo}} \cdot V \cdot d)\) while exploiting action-specific joint correlations in the adaptive adjacency matrices for dynamic neighbor selection. In addition, TATF dispenses with explicit Query–Key–Value projections and softmax normalization, yielding a lighter and numerically more stable design, and selectively enhances only low-frequency components so that high-frequency details crucial for fine-grained temporal variations are preserved.

\subsubsection{Multi-Scale Fusion and Classification}

We aggregate the fused features from all decomposition levels: \(X_{\text{agg}} = \sum_{j=1}^{J-1} X^{(j)}_{\text{fused}}\), where \(J = \lfloor \log_2 T \rfloor\) is the total number of decomposition levels. The scaling coefficient from the final level \(S^{(J)}\) is only upsampled without interaction and included in the aggregation. Subsequently, we apply Conv1d, BN, and SN to the aggregated features, followed by global average pooling to yield the multi-level output \(\hat{X}\).

The resulting feature \(\hat{X}\) is then fused with the SN-processed output of the final backbone layer \(X^{(L)}\), and the fused representation is sent to a fully connected layer for classification:

\begin{equation}
\hat{y} = \operatorname{FC}(\operatorname{GAP}(\operatorname{SN}(X^{(L)})) + \beta \hat{X})
\end{equation}
where \(\beta\) is a learnable parameter balancing the contribution of temporal-spatial and time-frequency features, and cross-entropy loss is used for optimization during training. This design enables the network to leverage both spatial topology-aware attention and multi-scale time-frequency representations for comprehensive action recognition.

\section{Experiments}
To demonstrate the effectiveness of Signal-SGN++, we conduct skeleton-based action recognition experiments on three large-scale datasets. We compare our model against strong ANN and SNN baselines, and perform ablation studies to evaluate the contribution of each module. The following sections cover the experimental setup in detail, including dataset settings, hyperparameter configurations, comparisons with state-of-the-art methods, and ablation studies.
\subsection{Dataset}

NTU RGB+D \cite{shahroudy2016ntu} is a large-scale benchmark dataset for 3D human action recognition, consisting of 56,880 skeleton-based video samples from 60 action classes. Each sample contains 3D coordinates of 25 human body joints captured by the Kinect v2 sensor. We follow the standard evaluation protocols, including cross-subject (Xs)—where training and testing subjects do not overlap—and cross-view (Xv)—where training and testing samples are captured from different camera angles. NTU-RGB+D 120 \cite{liu2019ntu} extends NTU 60 to 120 action categories and 114,480 video samples, with more subjects and greater view diversity. Evaluation is performed under cross-subject (Xs) and cross-setup (Xt) protocols, where the latter refers to training and testing samples being collected under different environmental and camera configurations, posing a greater challenge for model generalization. NW-UCLA \cite{wang2014cross} contains 10 action classes and 1,494 samples captured from three different viewpoints. Each sample includes 3D skeletons with 20 joints. The dataset is designed to test model robustness under cross-view scenarios.
\subsection{Model Implementation and Configuration}
Our model is implemented using PyTorch\footnote{\url{https://github.com/rwightman/pytorch-image-models}} and SpikingJelly\footnote{\url{https://github.com/fangwei123456/spikingjelly}}, and all experiments are conducted on four NVIDIA V100 GPUs. The backbone network consists of 4 layers with input and output channel sizes of 3 and 64, 64 and 64, 64 and 128, 128 and 256. We use SGD optimizer with a learning rate of 0.1, decay rate of 0.1 at step 110, batch size of 64, training for 150 epochs with a dropout rate of 0.3, weight decay of 0.0001, and auxiliary loss weight $\omega$ set to 0.03.

\begin{table*}[t]
\caption{Comparison of Signal-SGN++ with prior ANN- and SNN-based models on NTU RGB+D, NTU RGB+D 120, and NW-UCLA datasets. We report classification accuracy (\%), model size (Param.), computational cost (FLOPs and SOPs), and energy consumption (Power).}
\label{tab:full_comparison}
\centering
\begin{tabularx}{\textwidth}{X c c c c c c c c c}
\toprule
Model (ANN/SNN) & Param. & FLOPs& SOPs & Power
& \multicolumn{2}{c}{NTU RGB+D} 
& \multicolumn{2}{c}{NTU RGB+D 120} 
& NW-UCLA\\

& (M)& (G) &(G)& (mJ)
& Xs(\%) & Xv(\%) &Xs(\%) &Xt(\%) & (\%) \\
\midrule
Part-aware LSTM \cite{du2015hierarchical}         & -     & -     & -     & -      & 62.9 & 70.3 & 25.5 & 26.3 & - \\
Spatio-Temporal LSTM \cite{liu2016stlstm}         & -     & -     & -     & -      & 61.7 & 75.5 & 55.7 & 57.9 & - \\
ST-GCN \cite{Yan2018Spatial}                      & 3.10  & 3.48  & -     & 16.01  & 81.5 & 88.3 & 70.7 & 73.2 & - \\
2S-AGCN \cite{Shi2019Two}                         & 3.48  & 37.32 & -     & 182.87 & 88.5 & 95.1 & 82.5 & 84.3 & - \\
Shift-GCN \cite{Cheng2020Skeleton}                & -     & 10.00 & -     & 46.00  & 90.7 & 96.5 & 85.3 & 86.6 & 94.6 \\
MS-G3D \cite{Liu2020Disentangling}                & 3.19  & 48.88 & -     & 239.51 & 91.5 & 96.2 & 86.9 & 88.4 & - \\
CTR-GCN \cite{Chen2021Channel}                    & 1.46  & 7.88  & -     & 36.25  & 92.4 & 96.8 & 88.9 & 90.6 & 96.5 \\
Info-GCN \cite{chi2022infogcn}                    & 1.6 & 7.36 & -     & 33.86  & 92.3 & 96.7 & 89.2 & 90.7 & 96.5 \\
FR-Head \cite{zhou2023ldr}                    & 1.3  & 7.08  & -     & 32.57 & 92.8 & 96.8 & 89.5 & 90.9 & 96.8 \\

\midrule
Spikformer \cite{Zhou2022Spikformer}              & 4.78  & 24.07 & 1.69  & 2.17   & 73.9 & 80.1 & 61.7 & 63.7 & 85.4 \\
Spike-driven Transformer \cite{Yao2023SpikeTransformer}      & 4.77  & 23.50 & 1.57  & 1.93   & 73.4 & 80.6 & 62.3 & 64.1 & 83.4 \\
Spike-driven Transformer V2 \cite{Yao2024SpikeV2}  & 11.47 & 38.28 & 2.59  & 2.91   & 77.4 & 83.6 & 64.3 & 65.9 & 89.4 \\
Spiking Wavelet Transformer \cite{Fang2024SWformer}& 3.24 & 19.30 & 1.48  & 2.01   & 74.7 & 81.2 & 63.5 & 64.7 & 86.7 \\
MK-SGN \cite{Zheng2025MKSGN}                         & 2.17  & 7.84  & 0.68  & 0.614  & 78.5 & 85.6 & 67.8 & 69.5 & 92.3 \\
STAtten \cite{Lee2025STAtten}    & 3.19  & 25.48& 1.98 & 2.48 &72.8 & 79.7 & 60.3 & 61.7 & 86.9\\

Signal-SGN (Joint) \cite{2025signalsgn}                                 & 1.74  & 1.62  & 0.314 & 0.372  & 80.5 & 87.7 & 69.2 & 72.1 & 92.7 \\

Signal-SGN (Bone+Joint)\cite{2025signalsgn}                             & 1.74  & 3.24  & 0.628 & 0.744  & 82.5 & 89.2 & 71.3 & 74.2 & 93.1 \\

Signal-SGN (4 ensemble) \cite{2025signalsgn}                & 1.74&6.48 & 1.28&1.488& 86.1 & 93.1& 75.3& 77.9 & 95.9 \\
\rowcolor[gray]{0.93}
Signal-SGN++ (Joint)                                &  1.72 & 1.61 & 0.313  &0.369& 81.3 &88.4&70.5&73.2&93.5\\
\rowcolor[gray]{0.93}
Signal-SGN++(Bone+Joint)                          &  1.72 & 3.22 & 0.626 &0.738& 84.5 &90.1&72.7&75.6&93.7\\
\rowcolor[gray]{0.93}
Signal-SGN++(4 ensemble)               &  1.72 & 6.44 &1.252 &1.476& 87.2 &94.5&76.5&78.9&96.3\\
\bottomrule
\end{tabularx}
\footnotesize Param. represents the model size for the network itself. For ANN methods, FLOPs, power, and accuracy are reported under standard protocols. For SNN methods, SOPs and energy estimates are computed based on spike-driven inference.
\end{table*}

\subsection{Comparison with the State-of-the-Art}
Several SOTA methods adopt multi-stream fusion strategies for improved performance \cite{Cheng2020Skeleton, Chen2021Channel, chi2022infogcn}. To ensure a fair comparison, we employ the same four-stream fusion protocol in Signal-SGN++, incorporating joint, bone, joint motion, and bone motion modalities. We evaluate Signal-SGN++ against both ANN and SNN baselines on NTU RGB+D, NTU RGB+D 120, and NW-UCLA. Classification accuracy, model size, FLOPs, synaptic operations (SOPs), and estimated energy consumption are summarized in Table~\ref{tab:full_comparison}, where the energy consumption is computed using formulas adopted in several prior works\cite{Zhou2022Spikformer,Yao2023SpikeTransformer,Yao2024SpikeV2} and in a prior conference paper\cite{2025signalsgn}.The explicit formulas used for computing energy consumption are also included in the supplementary material.

As shown in Table\ref{tab:full_comparison}, ANN-based GCNs like CTR-GCN\cite{Chen2021Channel}  and FR-Head \cite{zhou2023ldr} achieve high accuracy, with FR-Head reaching 92.8\% on NTU RGB+D (Xs). However, this performance comes at a significant energy cost, with FR-Head consuming an estimated 32.57 mJ per inference. In contrast, our Signal-SGN++ (4-stream ensemble) achieves a competitive accuracy of 87.2\% on the same benchmark while consuming only a fraction of the energy. This represents a reduction of over an order of magnitude in energy consumption compared to SOTA ANN-based methods like FR-Head and Shift-GCN, highlighting the profound energy efficiency of our SNN-based approach. This demonstrates that Signal-SGN++ successfully bridges the gap between high performance and low power, making it a viable solution for deployment on energy-constrained edge devices.

When compared to other SNN-based models, Signal-SGN++ demonstrates a clear advantage in both accuracy and efficiency. On the challenging NTU-120 (Xs) benchmark, our 4-stream model achieves 76.5\% accuracy, surpassing the previous best SNN model, MK-SGN \cite{Zheng2025MKSGN}     (67.8\%), by a significant margin of 8.7\%. It also outperforms other SNN-Transformer models like Spike-driven Transformer V2 \cite{Yao2024SpikeV2} (64.3\%) and Spiking Wavelet Transformer \cite{Fang2024SWformer} (63.5\%) by over 12\%. Notably, the single-stream (Joint) model already outperforms most other SNN models while maintaining an extremely low energy footprint of just 0.369 mJ. This superior performance is attributable to the synergistic integration of the TSSA and MWTF modules, which enable more effective modeling of dynamic topologies and time-frequency features than prior SNN-based approaches. The results confirm that Signal-SGN++ sets a new state-of-the-art for SNN-based skeleton action recognition.
Signal-SGN++ consistently outperforms Signal-SGN~\cite{2025signalsgn} across all datasets and protocols. For instance, the 4-stream ensemble model shows a 1.1\% improvement on NTU RGB+D (Xs) and a 1.2\% improvement on NTU-120 (Xs). This validates the effectiveness of the newly introduced TSSA and MWTF modules. By achieving a compelling balance of high accuracy and ultra-low energy consumption, Signal-SGN++ establishes itself as a leading framework for efficient and robust skeleton-based action recognition.
\subsection{Ablation Studies}
To evaluate the contributions of the primary components in Signal-SGN++, we conducted ablation studies on the NTU-RGB+D (Xs) dataset based on joint data, with all sequences resized to 16 frames. 

\begin{table*}[t]
\centering
\caption{Ablation on NTU-RGB+D (Xs) following the pipeline \textbf{3D C.N. $\rightarrow$ 1D-SGC $\rightarrow$ Attention $\rightarrow$ FSC $\rightarrow$ MWTF}. 
We report parameters, FLOPs/SOPs per sample, and Top-1 accuracy under the same training protocol. 
\textbf{Gain} is computed vs. \textbf{B1} (1D-SGC w/ 3D C.N.).}
\label{table:ablation_journal_final}
\renewcommand{\arraystretch}{1.2}
\setlength{\tabcolsep}{4pt}
\begin{tabular}{c|c|c|c|c|c|ccc|cc}
\hline
\multirow{2}{*}{\textbf{ID}} & \multirow{2}{*}{\textbf{3D C.N.}} & \multirow{2}{*}{\textbf{Backbone}} & \multirow{2}{*}{\textbf{Attention}} & \multirow{2}{*}{\textbf{FSC}} & \multirow{2}{*}{\textbf{MWTF}} & \multicolumn{3}{c|}{\textbf{Resources}} & \multicolumn{2}{c}{\textbf{Result}} \\
\cline{7-11}
& & & & & & \textbf{Param.(M)} & \textbf{FLOPs (G)} & \textbf{SOPs (G)} & \textbf{Top-1 (\%)} & \textbf{Gain} \\
\hline
\multicolumn{11}{c}{\textit{External Baseline}} \\
\hline
A0 & -- & Spikformer~\cite{Zhou2022Spikformer} & -- & -- & -- & 4.78 & 24.07 & 1.69 & 73.9 & -- \\
\hline
\multicolumn{11}{c}{\textit{Stage I: 1D-SGC Backbone (no attention/FSC/MWTF), 3D C.N. ablation}} \\
\hline
B0 & --          & 1D-SGC & -- & -- & -- & 0.744 & 0.789 & 0.205 & 65.0& \textit{--} \\
B1 & \checkmark  & 1D-SGC & -- & -- & -- & 0.744 & 0.789 & 0.205 & 65.7 & -- \\
\hline
\multicolumn{11}{c}{\textit{Stage II: Attention on 1D-SGC (no FSC/MWTF), 3D C.N. enabled}} \\
\hline
B2 & \checkmark & 1D-SGC & SSA~\cite{Zhou2022Spikformer} & -- & -- & 1.03 & 1.12 & 0.252 & 69.7 & +4.0 \\
B3 & \checkmark & 1D-SGC & SDA~\cite{Yao2023SpikeTransformer} & -- & -- & 1.04 & 1.17 & 0.264 & 70.3 & +4.6 \\
B4 & \checkmark & 1D-SGC & STAttn~\cite{Lee2025STAtten} & -- & -- & 1.01 & 1.10 & 0.271 & 69.9 & +4.2 \\
B5 & \checkmark & 1D-SGC & FATM~\cite{Fang2024SWformer} & -- & -- & 0.998 & 1.04 & 0.248 & 70.1 & +4.4 \\
\rowcolor[gray]{0.93}
B6 & \checkmark & 1D-SGC & \textbf{TSSA} & -- & -- & 1.02 & 1.11 & 0.251 & \textbf{71.5} & \textcolor{red}{+5.8} \\
\hline
\multicolumn{11}{c}{\textit{Stage III: +FSC on top of 1D-SGC + Attention (no MWTF), 3D C.N. enabled}} \\
\hline
C1 & \checkmark & 1D-SGC & SSA & \checkmark & -- & 1.65 & 1.60 & 0.312 & 78.3 & +12.6 \\
C2 & \checkmark & 1D-SGC & \textbf{TSSA} & \checkmark & -- & 1.64 & 1.59 & 0.303 & \textbf{79.9} & \textcolor{red}{+14.2} \\
\hline
\multicolumn{11}{c}{\textit{Stage IV: +MWTF Fusion (built on 1D-SGC + \textbf{Attention} + FSC), 3D C.N. enabled}} \\
\hline
D1 & \checkmark & 1D-SGC & SSA & \checkmark & SCA & 1.74 & 1.62 & 0.314 & 80.5 & +14.8 \\
D2 & \checkmark & 1D-SGC & \textbf{TSSA} & \checkmark & SCA & 1.73 & 1.61 & 0.311 & 81.2 & +15.5 \\
\rowcolor[gray]{0.93}
D3 & \checkmark & 1D-SGC & \textbf{TSSA} & \checkmark & \textbf{TATF} & 1.72 & 1.61 & 0.313 & \textbf{82.5} & \textcolor{red}{+16.8} \\
\hline
\end{tabular}

\vspace{2pt}
\footnotesize{
Notes: (1) All numbers are under 16-frame input, X-subject protocol; hyper-parameters are fixed across rows unless stated. 
(2) FLOPs/SOPs are per-sample. Energy (mJ/sample) is reported in Tab.~\ref{tab:full_comparison}. 
(3) Row \textbf{B0} disables 3D C.N. for an apples-to-apples comparison; only Top-1 should change, while parameters and computation remain identical to \textbf{B1}.
}
\end{table*}
Stage I isolates the effect of 3D C.N. on the plain 1D-SGC backbone. Enabling 3D C.N. improves Top-1 from 65.0\% (B0) to 65.7\% (B1) without changing parameters or computation, indicating that normalizing the 3D coordinates provides a slightly more stable spatial prior before attention and frequency modeling.

Stage II compares different attention modules on top of 1D-SGC with 3D C.N. enabled and without FSC/MWTF. Under closely matched budgets (about 1.0--1.04 M parameters, 1.10--1.17 G FLOPs, and 0.248--0.271 G SOPs), TSSA achieves the best accuracy of 71.5\% (B6), outperforming SSA, SDA, STAttn, and FATM. Relative to the backbone B1, TSSA yields a +5.8-point gain for only moderate increases in parameters/FLOPs/SOPs, showing that topology-aware neighbor selection with shift augmentation converts a modest compute overhead into a substantial recognition benefit.

Stage III adds frequency-domain spiking convolution (FSC) on top of the attention modules. FSC brings a large performance jump: SSA increases from 69.7\% to 78.3\% (B2$\rightarrow$C1), and TSSA from 71.5\% to 79.9\% (B6$\rightarrow$C2). Compared with B1, the TSSA+FSC model C2 gains +14.2 points with about 2.2$\times$ parameters, 2.0$\times$ FLOPs, and 1.5$\times$ SOPs, indicating that modeling cross-joint frequency structure is a principal source of accuracy improvements for spiking skeleton recognition.

Stage IV introduces multi-scale wavelet-based fusion on top of 1D-SGC + Attention + FSC. Along the TSSA path, SCA raises accuracy from 79.9\% (C2) to 81.2\% (D2), and replacing SCA with TATF further improves it to 82.5\% (D3). The transition from C2 to D3 incurs only small overheads (about +5\% parameters, +1\% FLOPs, +3\% SOPs) yet adds a +2.6-point gain; overall, D3 achieves +16.8 points over B1. The SSA path with fusion (D1, 80.5\%) remains consistently behind the TSSA path, confirming that TSSA, FSC, and MWTF---especially TATF---form a complementary hierarchy: 3D C.N. offers a small but stable boost, TSSA is the most effective attention choice under comparable budgets, FSC provides the dominant leap, and MWTF/TATF delivers an additional reliable margin at minimal cost.

\begin{table}[t]
\centering
\caption{Fine-grained ablation of TSSA on NTU-RGB+D (Xs) under the 1D-SGC + 3D C.N. backbone. 
All models use 16-frame input with fixed feature dimensions. 
Gain is computed w.r.t.\ SSA.}
\label{tab:tssa_fine_ablation}
\begin{tabular}{l|c|c|c|cc}
\hline
Variant                     & $k$ & Shift     & FLOPs(G) & Top-1(\%) & Gain \\
\hline
SSA                         & --  & --        & 1.12 & 69.7 & -- \\[1pt]

$\mathcal{Q}_t^{(0)}$ only  &  8  & roll$\pm$1 & 1.11 & 70.2 & +0.5 \\
$\mathcal{A}$ only          &  8  & roll$\pm$1 & 1.11 & 69.9 & +0.2 \\
Static topology             &  8  & roll$\pm$1 & 1.11 & 70.5 & +0.8 \\[1pt]

TSSA, $k=4$                 &  4  & roll$\pm$1 & 1.06 & 70.8 & +1.1 \\
\rowcolor[gray]{0.93}
TSSA, $k=8$                 &  8  & roll$\pm$1 & 1.11 & \textbf{71.5} & \textbf{+1.8} \\
TSSA, $k=12$                & 12  & roll$\pm$1 & 1.19 & 71.6 & +1.9 \\[1pt]

TSSA no shift               &  8  & none       & 1.11 & 70.4 & +0.7 \\
TSSA roll+1                 &  8  & roll+1     & 1.11 & 71.0 & +1.3 \\
TSSA roll$-1$               &  8  & roll$-1$   & 1.11 & 70.9 & +1.2 \\
\hline
\end{tabular}
\end{table}

We further conduct a fine-grained ablation of TSSA under the same 1D-SGC + 3D C.N. backbone on NTU-RGB+D (Xs), as summarized in Table \ref{tab:tssa_fine_ablation}. It shows that neither using only similarity $\mathcal{Q}$, only topology $\mathcal{A}$, nor a static non-learnable graph is sufficient, all performing notably worse than the fused and learnable TSSA design. Varying the sparsity level further indicates that $k=8$ provides the best balance between accuracy and computation: smaller $k$ underfits due to insufficient neighbors, whereas larger $k$ increases FLOPs with marginal gains. Finally, evaluating the shift templates demonstrates that the bidirectional cyclic shift (roll$\pm$1) delivers the highest accuracy without additional cost, while removing the shift or using a single-direction roll leads to clear degradation. These results collectively validate the effectiveness of TSSA’s learnable fusion, moderate sparsity, and lightweight shift mechanism. 

\begin{table*}[t]
\centering
\caption{Comparison of MWTF/TATF configurations on NTU-RGB+D (Xs) under the 1D-SGC + TSSA + FSC backbone.
We vary MWTF decomposition parameters $(m,J)$, the topology neighborhood size $k_{\text{topo}}$, the wavelet bands with topology, and the topology source.
All models use 16-frame input.}
\label{tab:tatf_fine_ablation}

\begin{tabular}{l|c|c|c|c|c|ccc}
\hline
Setting & $m$ & $J$ & $k_{\text{topo}}$ & Topology bands & Topology source 
        & Param.(M) & FLOPs(G) & Top-1(\%) \\
\hline
\multicolumn{9}{c}{\textit{(A) Effect of wavelet bands and topology}} \\
\hline
MWTF only (S\&D, no TATF) & 8 & 3 & 0 & $S^{(j)}, D^{(j)}$ & --                 & 1.70 & 1.57 & 81.0 \\
MWTF S-only (no TATF)     & 8 & 3 & 0 & $S^{(j)}$          & --                 & 1.70 & 1.54 & 80.5 \\
MWTF D-only (no TATF)     & 8 & 3 & 0 & $D^{(j)}$          & --                 & 1.70 & 1.54 & 80.2 \\
\rowcolor[gray]{0.93}
TATF low-only (ours)      & 8 & 3 & 6 & $S^{(j)}$          & learned-$\mathcal{A}$ & 1.72 & 1.60 & \textbf{82.5} \\
TATF all-bands            & 8 & 3 & 6 & $S^{(j)}, D^{(j)}$ & learned-$\mathcal{A}$ & 1.72 & 1.63 & 81.8 \\
\hline

\multicolumn{9}{c}{\textit{(B) Effect of topology source (fixed $m{=}8, J{=}3$, low-only)}} \\
\hline
Static-$S$              & 8 & 3 & 6 & low-frequency $S^{(j)}$    & static skeleton       & 1.72 & 1.60 & 81.4 \\
Random-topology         & 8 & 3 & 6 & low-frequency $S^{(j)}$    & random sparse         & 1.72 & 1.60 & 80.7 \\
\hline

\multicolumn{9}{c}{\textit{(C) Effect of $k_{\text{topo}}$ (fixed $m{=}8, J{=}3$, low-only, learned-$\mathcal{A}$)}} \\
\hline
$k_{\text{topo}} = 2$   & 8 & 3 & 2 & low-frequency $S^{(j)}$ & learned-$\mathcal{A}$ & 1.72 & 1.58 & 81.3 \\
$k_{\text{topo}} = 4$   & 8 & 3 & 4 & low-frequency $S^{(j)}$ & learned-$\mathcal{A}$ & 1.72 & 1.59 & 82.0 \\
$k_{\text{topo}} = 8$   & 8 & 3 & 8 & low-frequency $S^{(j)}$ & learned-$\mathcal{A}$ & 1.72 & 1.61 & 82.3 \\
\hline

\multicolumn{9}{c}{\textit{(D) Effect of $m$ (fixed $J{=}3$, $k_{\text{topo}}{=}6$, low-only, learned-$\mathcal{A}$)}} \\
\hline
$m = 4$                 & 4  & 3 & 6 & low-frequency $S^{(j)}$ & learned-$\mathcal{A}$ & 1.70 & 1.52 & 82.0 \\
$m = 12$                & 12 & 3 & 6 & low-frequency $S^{(j)}$ & learned-$\mathcal{A}$ & 1.75 & 1.70 & 82.4 \\
\hline
\end{tabular}

\end{table*}
The fine-grained analysis of MWTF/TATF in Table~\ref{tab:tatf_fine_ablation} reveals several consistent trends. First, decomposing the input into both low- and high-frequency components (S\&D) is more effective than using only $S^{(j)}$ or only $D^{(j)}$, but applying topology to all bands slightly harms performance, indicating that structural aggregation is most beneficial in the smooth, low-frequency domain where global motion patterns dominate. Accordingly, the TATF low-only design achieves the best overall accuracy (82.5\%). Second, the source of topology plays a critical role: replacing the learned affinity with a static skeleton or random sparse graph consistently reduces accuracy, confirming that reusing the action-specific topology extracted by the final TSSA layer is essential for effective cross-frequency fusion. Third, varying the neighborhood size shows that very small $k_{\text{topo}}$ underutilizes meaningful neighbors, while excessively large values incur unnecessary aggregation; a moderate choice ($k_{\text{topo}}=6$) offers the best trade-off. Finally, the MWTF decomposition width $m$ exhibits a similar pattern: too small a bandwidth limits representational capacity, whereas overly large values increase cost with marginal benefit, with $m=8$ performing best under fixed $J=3$. These results jointly validate the design choices of TATF, demonstrating that selective low-frequency topology, learned structural priors, and moderate fusion bandwidth are key to the strong performance of the proposed framework.

\subsection{Visualization and Analysis}

To complement the quantitative results and provide a deeper, qualitative validation of our proposed Signal-SGN++ framework, we conduct a series of visualization experiments. This analysis aims to elucidate the inner workings of our key modules and demonstrate their effectiveness in capturing dynamic topological structures and time-frequency characteristics. The following subsections are organized to provide a multi-faceted view of the model's behavior, from high-level adaptive patterns to the micro-mechanisms of our proposed attention and fusion units.

\subsubsection{Analysis of Learned Model-Level Dynamics}

We visualize three complementary aspects to probe the hierarchy in Signal-SGN++. 

\begin{figure}[h]
    \centering
    \includegraphics[width=\linewidth]{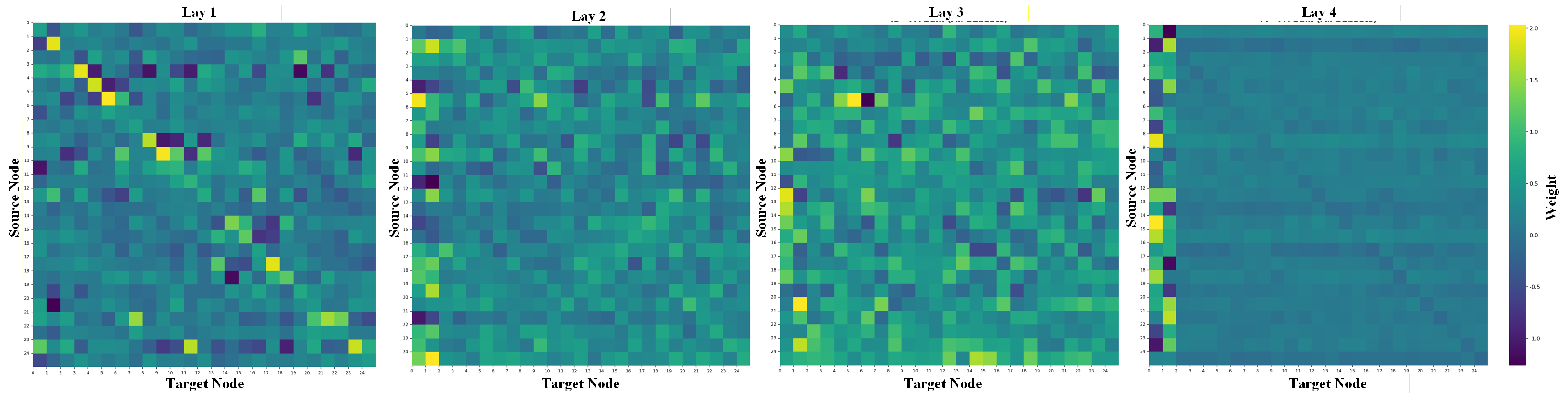}
    \caption{Layer-wise and accumulated topology score matrices. Off-diagonal mass increases with depth, revealing emergent nonlocal couplings and action-specific coordination.}
    \label{fig:pa-sum}
\end{figure}

(i) For each layer, we plot the topology score matrix and its cumulative form, as shown in Figure\ref{fig:pa-sum}. In the early layers, the matrix primarily emphasizes \textbf{local physical adjacencies}, focusing on nearby joints. As the layers increase, off-diagonal connections become more prominent, indicating the model’s ability to capture long-range, nonlocal interactions across body parts. This shift in the topology score matrix highlights the model’s hierarchical learning process, where it gradually uncovers more complex, action-dependent joint dependencies, ultimately enabling the model to capture global coordination crucial for action recognition.

\begin{figure}[h]
    \centering
    \includegraphics[width=\linewidth]{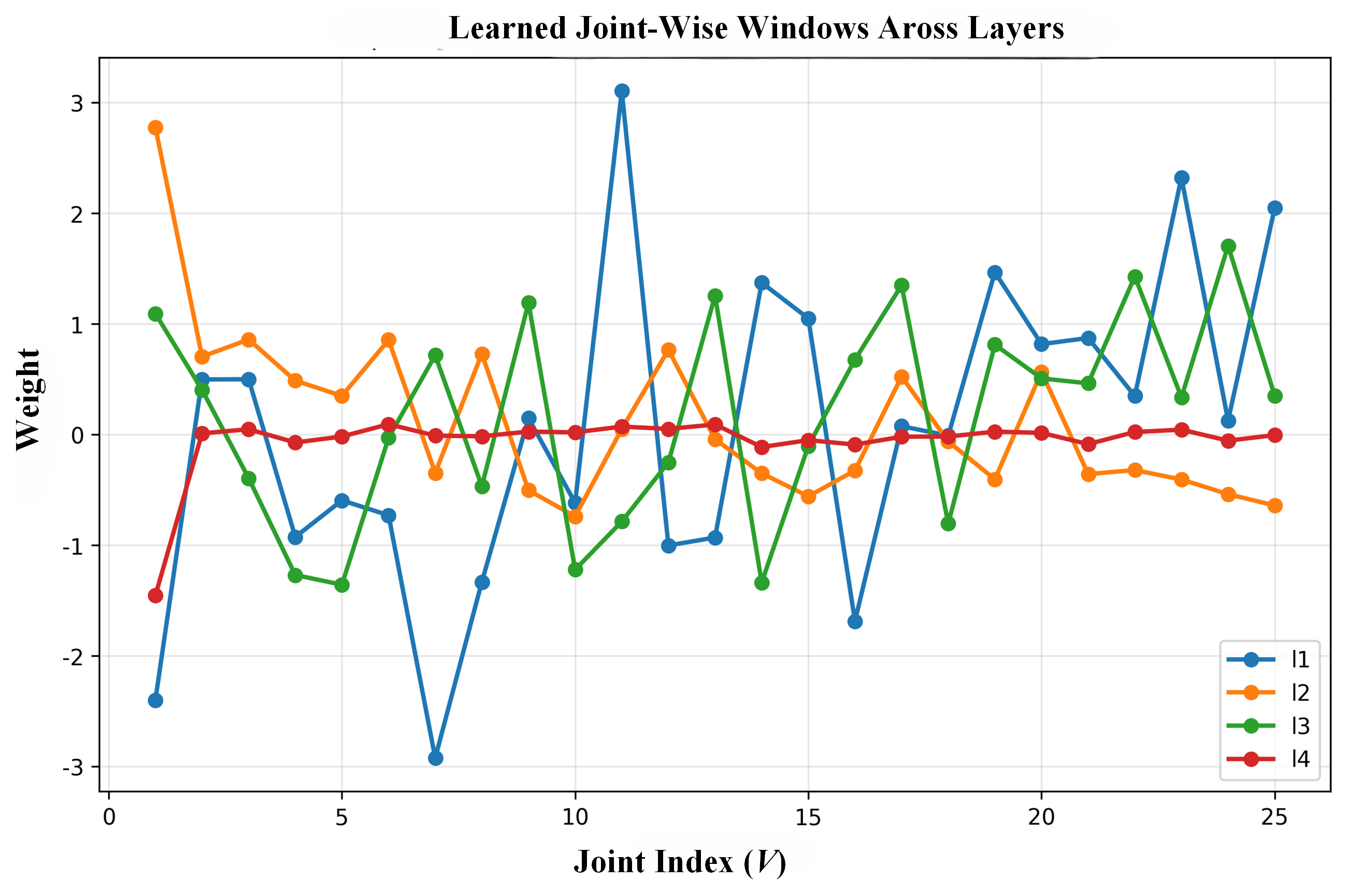}
    \caption{Learned \textbf{joint-wise} windows $w_V$ across layers (L1--L4). Shallow layers emphasize distal effectors; deeper layers smooth around the trunk, enlarging the effective receptive field.}
    \label{fig:joint-window}
\end{figure}

(ii) We further report the \textbf{joint-wise} learnable window $w_V$ from Layer1(L1) to Layer4(L4) (Fig.~\ref{fig:joint-window}), which gates informative joints along the joint axis before the frequency transform. 
In early layers (L1 and L2), the attention weights are more dispersed, with a higher emphasis on distal effectors such as hands and feet, reflecting the local fine-grained movements at the initiation of actions. As the layers deepen, the learned window functions smoothen, placing more focus on the torso and core body regions. This shift in focus suggests that the model transitions from capturing localized, motion-specific details to understanding more global, coordinated actions that involve the central body regions, which is crucial for recognizing complex actions.

\begin{figure}[t]
    \centering
    \includegraphics[width=\linewidth]{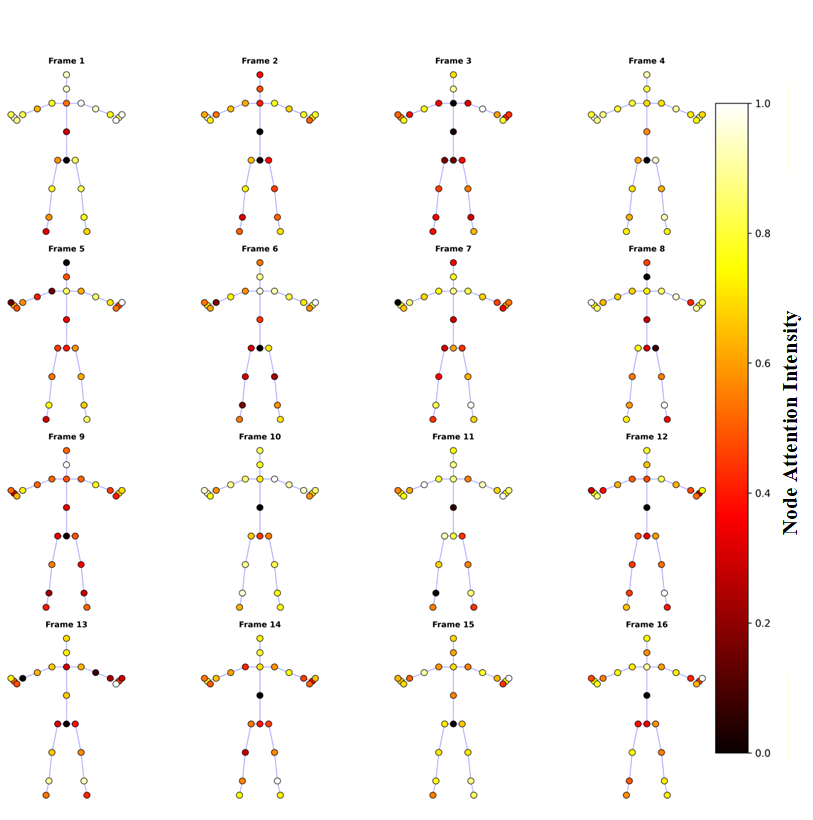}
    \caption{Spatio--temporal attention visualization for the ``clapping'' class. Energy concentrates on bilateral hand--forearm chains and aligns with upper-torso nodes near impact frames; background frames remain suppressed.}
    \label{fig:clap-attn}
\end{figure}

(iii) Finally, we present the spatio--temporal responses of the proposed TSSA mechanism, highlighting how it adapts to different action phases in the ``clapping'' class (Fig.~\ref{fig:clap-attn}). 
During the initial phase of the action (Frames 1-4), attention is concentrated on the hands and forearms, which are pivotal in the start of the clapping motion. As the action progresses (Frames 5-8), attention spreads to the upper torso, reflecting the synchronization between the arms and torso. In the later phases (Frames 9-16), attention stabilizes on the hands and forearms, with the torso maintaining a steady, less active role. These results demonstrate the ability of TSSA to selectively focus on the critical body parts at different time frames, adapting its attention to the dynamic motion of the action and significantly enhancing the model's spatio-temporal representation of complex actions.

\subsubsection{Visualization and Analysis of the TSSA Module}
This section visually analyzes the TSSA module to validate its mechanisms in skeleton-based action recognition. We show how TSSA selects joint neighbors, performs feature shifts, and fuses topological structure with feature similarity, demonstrating its ability.

\begin{figure}[t]
    \centering
    \includegraphics[width=\linewidth]{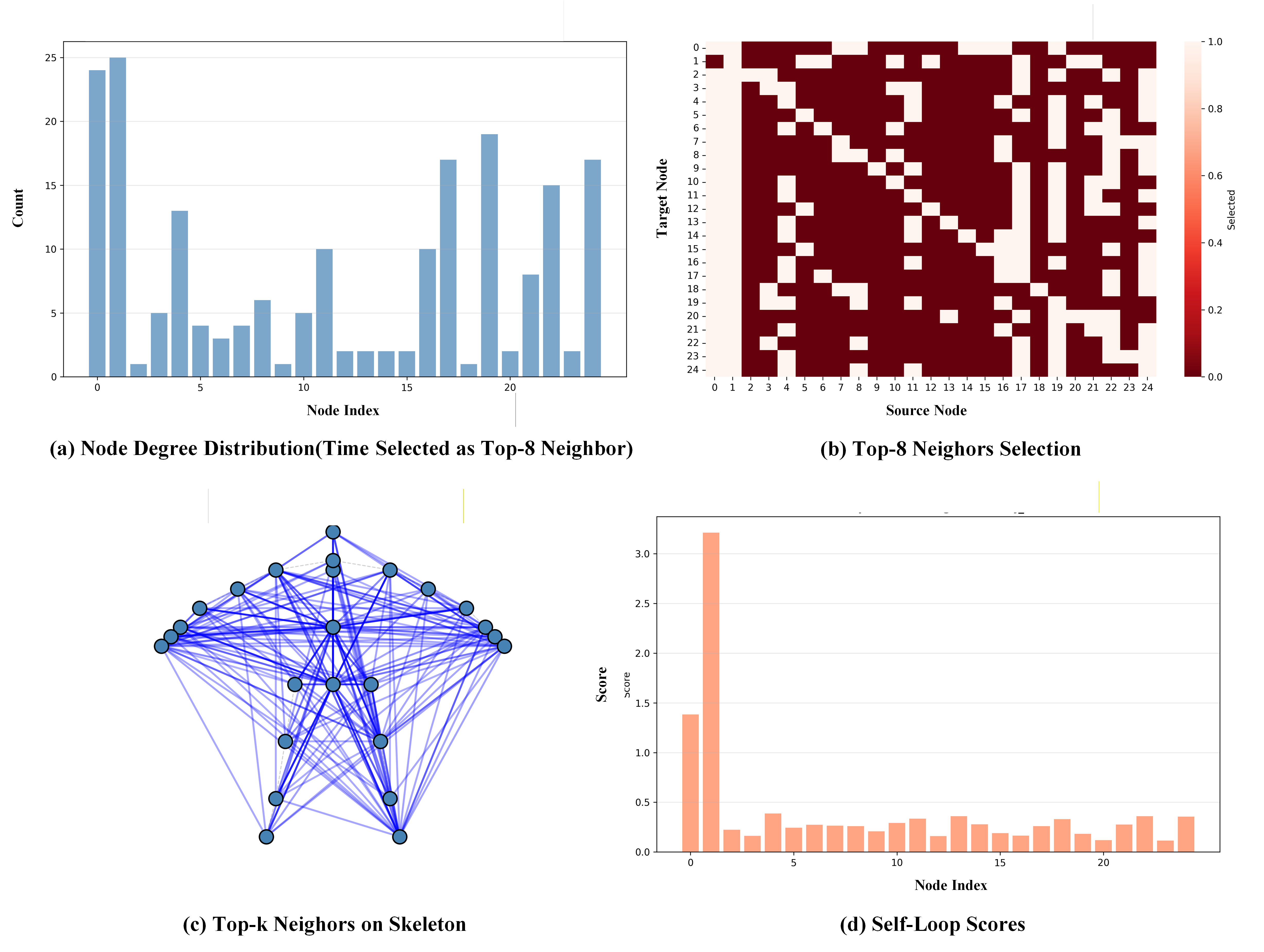}
    \caption{This figure visualizes key processes in the TSSA module, including node degree distribution, Top-8 neighbor selection, Top-k neighbors on the skeleton, and self-loop scores. These visualizations show how the module dynamically selects joints and adjusts attention for action recognition.}
    \label{fig:qkSIMILTY}
\end{figure}

\begin{figure}[t]
    \centering
    \includegraphics[width=\linewidth]{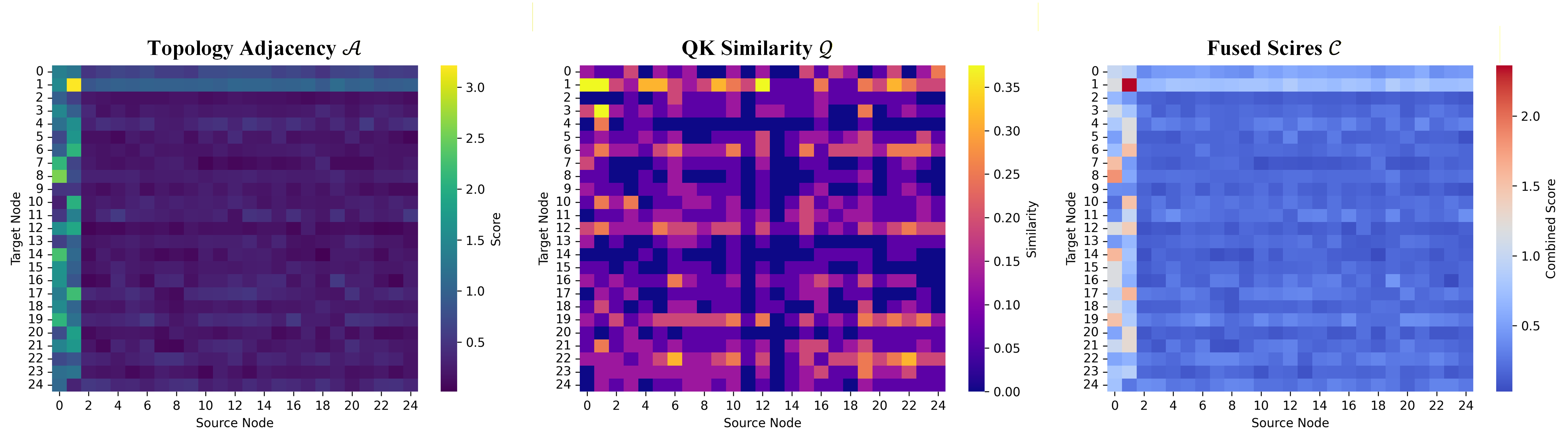}
    \caption{Visualization of the TSSA module at layer l4 and time step t=8 for the Clapping action, showing the fusion of topological adjacency, QK similarity, and fused scores. This demonstrates how the module integrates topological structure and feature similarity for joint selection, optimizing performance and reducing computational complexity.}
    \label{fig:clap-simlarity}
\end{figure}

We visualize TSSA on the \emph{Clapping} class at layer 4 and time step $t{=}8$ to validate its mechanisms. In Fig.~\ref{fig:qkSIMILTY}, the node-degree distribution shows that spine and shoulder joints are most frequently chosen as top-8 neighbors, reflecting their central role and higher dynamic relevance. The top-8 selection matrix indicates sparse, similarity-driven neighbor choice that prioritizes joints engaged in key motion while ignoring irrelevant ones. The top-$k$ overlay on the skeleton further reveals that locally active joints (e.g., hands) attract fewer neighbors, whereas more global movements (e.g., arm swings) induce broader connectivity. Self-loop scores remain high for stable joints such as the spine, preserving joint identity when motion is limited. Fig.~\ref{fig:clap-simlarity} contrasts the topological adjacency, QK similarity, and the fused scores. Topology highlights structurally central couplings (brighter entries around torso–shoulder regions), while the QK map emphasizes feature co-activation among coordinated parts (hands–arms). The fused scores integrate these cues, concentrating attention on action-critical joints (hands/shoulders) and suppressing less informative ones, showing that TSSA effectively balances structural priors with data-driven similarity for joint selection.

\begin{figure}[t]
    \centering
    \includegraphics[width=\linewidth]{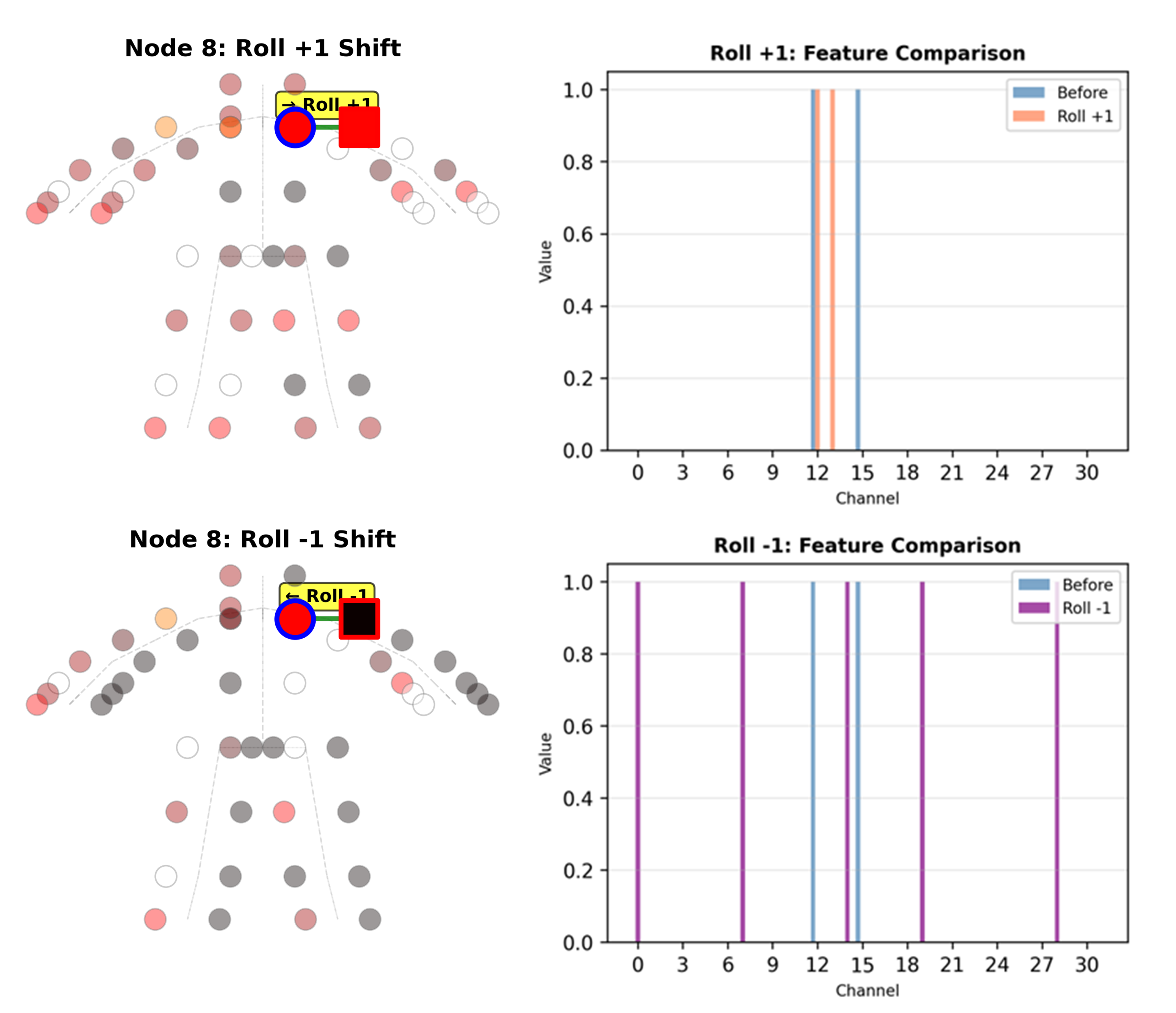}
    \caption{Visualization of the roll+1 and roll-1 feature shift operations applied to node 8 at time step t=8 during the Clapping action. The left side shows the joint features before and after the shifts, while the right side presents the feature comparison across channels.}
    \label{fig:tssa-roll}
\end{figure}

Finally, Fig.~\ref{fig:tssa-roll} examines the channel-shift templates. Applying right/left cyclic shifts (\textit{roll+1}/\textit{roll-1}) reassigns activations across channels without altering their magnitudes, acting as a lightweight permutation rather than an additional multiplication. This provides a cheap surrogate for temporal mixing: information from the top-$k$ neighbors is redistributed across feature channels, enriching the aggregated representation while preserving energy efficiency and numerical stability.

\subsubsection{Visualization and Analysis of the MWTF Module}
\begin{figure}[h]
    \centering
    \includegraphics[width=\linewidth]{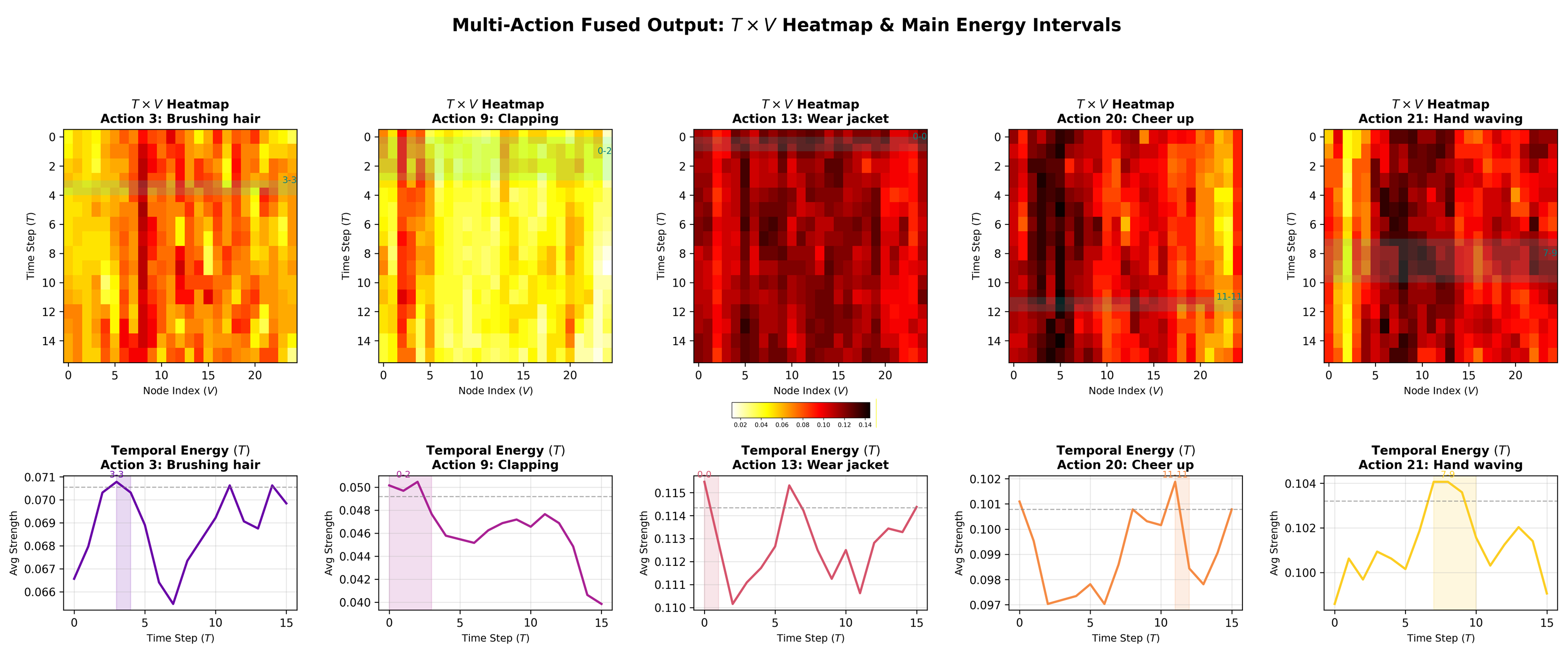}
    \caption{MWTF fused spatio–temporal responses for multiple actions: \(T \times V\) heatmaps (top) and temporal energy curves with main energy intervals (bottom).}
    \label{fig:MWTF_STresonse}
\end{figure}
To further verify that the proposed MWTF--TATF pipeline indeed focuses on discriminative spatio--temporal regions while suppressing irrelevant frames, we conduct a qualitative analysis of the fused features, as illustrated in Fig.~\ref{fig:MWTF_STresonse}. Specifically, we first average the MWTF output over channels and visualize the resulting \(T \times V\) maps as heatmaps (top row), and then average over joints to obtain temporal energy curves with automatically detected main energy intervals (bottom row). The heatmaps show that the fused responses form compact high-activation regions that align well with the semantics of each action, rather than being uniformly distributed across all joints and frames. Meanwhile, the temporal energy curves exhibit contiguous high-energy segments that correspond to the dominant motion phases of each action, with low responses outside these intervals. These observations support our claim that the proposed fusion mechanism adaptively emphasizes key spatio--temporal patterns and effectively filters out redundant or less informative frames.
\section{Conclusion}
In this article, we proposed Signal-SGN++, a topology-enhanced time–frequency spiking graph network for skeleton-based action recognition. Building on a dual-branch spiking backbone with 1D Spiking Graph Convolution (1D-SGC) and Frequency Spiking Convolution (FSC), Signal-SGN++ jointly models spatial topology, temporal dynamics, and frequency structure under an energy-efficient spiking paradigm. The proposed Topology-Shift Self-Attention (TSSA) enables adaptive, shift-augmented neighbor selection on learned skeletal graphs, while the Multi-Scale Wavelet Transform Fusion (MWTF) branch and the Topology-Aware Time–Frequency Fusion (TATF) unit perform topology-consistent fusion of multi-resolution temporal–frequency components. Extensive experiments on large-scale skeleton benchmarks show that Signal-SGN++ achieves a strong accuracy–efficiency trade-off, consistently outperforming prior SNN-based approaches and remaining competitive with state-of-the-art GCN-based models under substantially reduced computational and energy budgets; ablation and visualization results further confirm that topology-aware attention, frequency-domain spiking convolution, and topology-guided wavelet fusion allow the network to focus on discriminative joints and compact action-dependent temporal windows while suppressing redundant motion.

\section{Appendix}
In the appendix, we first provide a supplemental detailed description of Power Consumption, Legendre Polynomial-Based Filters, Algorithm and Supplementary Analysis.
\subsection{Power Consumption}
\label{app:Powerconsumption}
We assume that multiply-and-accumulate (MAC) and accumulate (AC) operations are implemented on the 45 nm technology node with $E_{\text{MAC}}$ = 4.6 pJ and $E_{\text{AC}}$ = 0.9 pJ. 
The  \(\mathrm{FLOPs++}\) for Signal-SGN and all reproduced models are calculated using FVCore\footnote{\url{https://github.com/facebookresearch/fvcore}}, providing accurate estimates for each model component. These FLOPs are then combined with the firing rate and spike time steps to compute SOPs as:
\begin{equation}\mathrm{SOPs}(l) = fr \times T \times \mathrm{FLOPs}(l),\end{equation} 
where $fr$ is firing rate, $T$ is the spike time steps, \(\operatorname{FLOPs}(l)\) refers to the floating-point operations of \(l\), defined as the number of MAC operations performed in $l$ block or layer, and $\operatorname{SOPs}$ are the number
of spike-based AC operations.

We estimate the theoretical energy consumption of
Signal-SGN++ according to (\cite{kundu2021spikethrift,hu2021advancing,kundu2021hiresnn,kim2021optimizing}) and the model's theoretical energy consumption is described as:
\begin{equation}
\begin{gathered}
E_{\text {Signal-SGN++}}=E_{\text{EAC}} \times \mathrm{FL}_{\mathrm{SNN} \text { Conv }}^i+ \\
E_{\text{AC}} \times\left[\left(\sum_{n={\hat{n}}}^N \mathrm{SOP}_{\mathrm{SNN} \text { Conv }}^n+\sum_{m=1}^M \mathrm{SOP}_{\mathrm{SNN} \text { FC }}^m\right.\right. \\
\left.\left.+\sum_{l=1}^L \mathrm{SOP}_{\mathrm{SNN} 
 \text{ TSSA }}^l+\sum_{f=1}^F \mathrm{SOP}_{\mathrm{SNN} \text{ FFT }}^f\right)+\mathrm{SOP}_{\mathrm{SNN} \text{ MWTF }}\right]
\end{gathered}
\end{equation}
where $\mathrm{FL}_{\mathrm{SNN} \text { Conv }}$ is the first layer to encode skeleton into spike-form. Then the SOPs
of $N$ SNN Conv layers, $M$ SNN Fully Connected Layer (FC), $L$ SSA layer , $F$ FFT layer and MWTF are added together are added together and
multiplied by $E_{\text{AC}}$.
The theoretical energy consumption for ANN and SNN models is calculated as follows:

For ANNs, the energy consumption of a block \(b\) is given by:
\begin{equation}
\text{Power}(b) = E_{\text{EAC}} \times \mathrm{FLOPs}(b).
\end{equation}

For SNNs, the energy consumption of a block \(b\) is given by:
\begin{equation}
\text{Power}(b) =E_{\text{AC}}\times \mathrm{SOPs}(b).
\end{equation}
\subsection{Legendre Polynomial-Based Filters}
\label{app:filter}
Legendre polynomials $P_k(x)$, defined over the interval $[-1,1]$, provide a robust orthogonal basis for constructing frequency-domain filters aimed at decomposing spike-like features in temporal signals. Their orthogonality condition ensures that different polynomial orders yield mutually independent basis functions.
The polynomials are generated recursively using the relation:
\begin{equation}
\begin{aligned}
& P_0(x) = 1, \quad P_1(x) = x, \\
(m+1) P_{m+1}(x) &= (2k+1)xP_k(x) - kP_{m-1}(x).
\end{aligned}
\end{equation}
Their orthogonality condition,as:
\begin{equation}
\int_{-1}^{1} P_m(x)P_n(x)\,dx = \frac{2}{2n+1}\delta_{mn}
\end{equation}
This property is particularly advantageous for designing a set of filters that can isolate specific frequency components with minimal interference.

To apply these polynomials to temporal-domain signals sampled over $[0,T]$, we first normalize the time index:
\begin{equation}
u = \frac{t}{T}, \quad u \in [0,1].
\end{equation}
We then map $[0,1]$ to $[-1,1]$:
\begin{equation}
x = 2u - 1.
\end{equation}
This transformation aligns the original time domain with the canonical domain of the Legendre polynomials, ensuring straightforward evaluation of $P_k(x)$ at the corresponding sampling points.

Within this framework, low-frequency (“low-pass”) filters are derived from lower-order Legendre polynomials, which vary slowly across $[-1,1]$. To ensure proper normalization, each polynomial is scaled by $\sqrt{2m+1}$. Introducing a half-period shift $\Delta = 0.5$ in the normalized time domain allows us to construct two distinct low-pass filters by evaluating the polynomials at slightly different points. Specifically, the low-pass filters are defined as:
\begin{equation}
\Lambda_0[m,t] = \sqrt{2m+1}\,P_k(2u-1),
\end{equation}
\begin{equation}
\Lambda_1[m,t] = \sqrt{2m+1}\,P_k(2(u+0.5)-1) = \sqrt{2m+1}\,P_m(2u).
\end{equation}
Here, $\Lambda_0[m,t]$ and $\Lambda_1[m,t]$ share the same polynomial family but are evaluated at $2u-1$ and $2u$, respectively, introducing a phase-like shift that distinguishes the two filters in their temporal and spectral responses.

For high-frequency (“high-pass”) filters, the ideal construction involves deriving them from multi-resolution analysis (MRA) conditions to ensure orthogonality and perfect reconstruction. Nonetheless, a representative scheme, consistent with the normalization above, is given by:
\begin{equation}
\Gamma_0[m,t] = \sqrt{2}\,\Lambda_0[m,t],
\end{equation}
\begin{equation}
\Gamma_1[m,t] = \sqrt{2}\,\Lambda_1[m,t].
\end{equation}

Taken together, these four filters $(\Lambda_0,\Lambda_1,\Gamma_0,\Gamma_1)$ can decompose the signal into distinct frequency bands, enabling a nuanced analysis of spike-like features. By carefully mapping temporal samples into the Legendre polynomial domain, employing a normalized and shifted set of low-pass filters, and incorporating a high-pass counterpart, this approach provides a stable, orthogonal, and frequency-selective analytical framework. This framework facilitates effective characterization, improved interpretability, and subsequent analyses of temporal-varying signals in a multi-resolution context.

\subsection{Algorithm}
This section presents the essential pseudocode and detail description underpinning the computational framework discussed in this work. 
\subsubsection{Spiking Neuron Dynamics Simulation}
\label{APPendiSND}
The Spiking Neuron Dynamics describes the core mechanisms governing the temporal dynamics of spiking neurons within the network, outlined in Algorithm \ref{alg:lft}. 
\begin{algorithm}
\caption{Spiking Neuron Dynamics Simulation}
\label{alg:lft}
\begin{algorithmic}[1]
\STATE \textbf{Input:} $G \in \mathbb{R}^{T \times D \times V}$
\STATE \textbf{Parameters:} $\tau$, $V_{\text{rest}}$, $R$, $V_{\text{th}}$
\STATE \textbf{Output:} $G_o \in \mathbb{R}^{T \times C \times V}$
\STATE Initialize $V(t)$ and $G_o$
\FOR{$t = 0$ to $T-1$}
    \STATE $G_i(t) \gets G[t, :, :]$
    \STATE $\tau \frac{dV(t)}{dt} \gets -\left(V(t) - V_{\text{rest}}\right) + R \cdot G_i(t)$
    \IF{$V(t) \geq V_{\text{th}}$}
        \STATE $G_o[t, :, :] \gets 1$
        \STATE $V(t) \gets V_{\text{rest}}$  \COMMENT{Reset voltage after spike}
    \ELSE
        \STATE $G_o[t, :, :] \gets 0$
    \ENDIF
\ENDFOR
\STATE \textbf{return} $G_o\in \mathbb{R}^{T \times D \times V}$
\end{algorithmic}
\end{algorithm}

The input, \( G \in \mathbb{R}^{T \times D \times V} \), represents the temporal sequence of feature maps for \( T \) time steps, where \( D \) and \( V \) denote the feature dimension and the number of vertices (e.g., joints in skeleton data), respectively. The algorithm computes the output \( G_o \in \mathbb{R}^{T \times C \times V} \), which represents the spike emissions over time, with \( C \) corresponding to the output channel dimension.

At each time step \( t \), the membrane potential \( V(t) \) is updated based on the input current \( G_i(t) \) and the parameters governing spiking neuron dynamics, including the membrane time constant \( \tau \), the resting potential \( V_{\text{rest}} \), the resistance \( R \), and the firing threshold \( V_{\text{th}} \). Specifically, the potential evolves according to the equation:
\begin{equation}
\tau \frac{dV(t)}{dt} = -\left(V(t) - V_{\text{rest}}\right) + R \cdot G_i(t),
\end{equation}
where \( G_i(t) = G[t, :, :] \) is the input tensor for time step \( t \). If the membrane potential \( V(t) \) exceeds the threshold \( V_{\text{th}} \), a spike is emitted (\( G_o[t, :, :] = 1 \)), and the potential is reset to \( V_{\text{rest}} \). Otherwise, no spike is generated (\( G_o[t, :, :] = 0 \)).

This mechanism enables the simulation of spike-based information transmission, where \( G_o \) captures the spiking behavior at each time step. The tensor \( G_o \) has the same temporal and spatial resolution as the input but reflects the binary spike activations, effectively encoding temporal dynamics into discrete spike events.

The iterative process ensures that spiking neurons dynamically respond to varying inputs while adhering to biologically inspired principles. Subsequent layers of spiking neurons follow similar dynamics, allowing the network to propagate spike-based representations through deeper processing stages. 
\subsubsection{Iterative Decomposition}
\label{APPendIterativeDecomposition}
    The Iterative Decomposition algorithm describes the process of decomposing spike-form features into their frequency domain representations. This hierarchical decomposition enables the separation of temporal dynamics into high-frequency detail coefficients and low-frequency scaling coefficients, which are critical for efficient feature extraction and multi-resolution analysis in SNNs.

\begin{algorithm}
\caption{Iterative Decomposition}
\label{alg:algorithm}
\begin{algorithmic}[1]
\STATE $\widetilde{X}_0 \gets \widetilde{X}$
\FOR{$j = 0$ to $J-1$}
    \STATE $X_{\text{even}}^{(j)} \gets \widetilde{X}_j[::2, :, :]$
    \STATE $X_{\text{odd}}^{(j)} \gets\widetilde{X}_j[1::2, :, :]$
    \STATE $D_j \gets \Gamma_0 \cdot X_{\text{even}}^{(j)} + \Gamma_1 \cdot X_{\text{odd}}^{(j)}$
    \STATE $S_j \gets \Lambda_0 \cdot X_{\text{even}}^{(j)} + \Lambda_1 \cdot X_{\text{odd}}^{(j)}$
    \STATE $\widetilde{X}_{j+1} \gets S_j$
\ENDFOR
\STATE $S_J \gets \widetilde{X}_J$
\STATE \textbf{return} $\{D_j\}_{j=0}^{J-1}$ and  $\{S_j\}_{j=0}^{J}$
\end{algorithmic}
\end{algorithm}

The iterative decomposition process begins with the extended spike-form feature tensor \( \widetilde{X}  \), which serves as the initial input. The tensor is initialized as \( X_0 \), providing the starting representation for decomposition across \( J \) levels.

At each decomposition level \( j \), the input tensor \( X_j \) is divided along its temporal dimension into even and odd indexed components, denoted as \( X_{\text{even}}^{(j)} \) and \( X_{\text{odd}}^{(j)} \), respectively. These components are independently processed to compute the detail coefficients \( D_j \) and scaling coefficients \( S_j \) as follows:
\begin{equation}
D_j = \Gamma_0 \cdot X_{\text{even}}^{(j)} + \Gamma_1 \cdot X_{\text{odd}}^{(j)},
\end{equation}
\begin{equation}
S_j = \Lambda_0 \cdot X_{\text{even}}^{(j)} + \Lambda_1 \cdot X_{\text{odd}}^{(j)}.
\end{equation}
where, \( \Gamma_0, \Gamma_1 \) are high-pass filter matrices used to extract the detail coefficients \( D_j \), which capture high-frequency temporal variations. Conversely, \( \Lambda_0, \Lambda_1 \) are low-pass filter matrices that compute the scaling coefficients \( S_j \), preserving the low-frequency temporal information.

After computing \( D_j \), the scaling coefficients \( S_j \) are propagated to the next level by setting \( X_{j+1} = S_j \). This iterative decomposition continues across all \( J \) levels, refining the temporal representation at each step. Upon completing the \( J \)-th level, the final scaling coefficients \( S_J \) are obtained. The process outputs the set of detail coefficients \( \{D_j\}_{j=0}^{J-1} \) and scaling coefficients \( \{S_j\}_{j=0}^J \), providing a hierarchical representation of the temporal dynamics. 

Upon completion, the algorithm outputs a set of detail coefficients \( \{D_j\}_{j=0}^{J-1} \) and the final scaling coefficients \( S_J \), collectively encapsulating the hierarchical temporal structure of the input. The complete iterative decomposition process is outlined in Algorithm \ref{alg:algorithm}, providing a systematic and computationally efficient method for processing spike-form feature.


\begin{thebibliography}{1}
\bibliographystyle{IEEEtran}
\bibitem{kong2022survey}
Y.~Kong and Y.~Fu,
``Human Action Recognition and Prediction: A Survey,''
\emph{International Journal of Computer Vision}, vol.~130, no.~5, pp.~1366--1401, 2022.
\bibitem{sun2022review}
Z.~Sun, Q.~Ke, H.~Rahmani, \emph{et~al.},
``Human Action Recognition from Various Data Modalities: A Review,''
\emph{IEEE Transactions on Pattern Analysis and Machine Intelligence}, vol.~45, no.~3, pp.~3200--3225, 2022.

\bibitem{Du2015Skeleton}
Y. Du, Y. Fu, and L. Wang, ``Skeleton based action recognition with convolutional neural network,'' in \emph{2015 23rd International Conference on Pattern Recognition (ICPR)}, 2015, pp. 579-583.

\bibitem{du2015hrnn}
Y.~Du, W.~Wang, and L.~Wang,
``Hierarchical Recurrent Neural Network for Skeleton Based Action Recognition,''
in \emph{Proceedings of the IEEE Conference on Computer Vision and Pattern Recognition (CVPR)}, 2015, pp.~1110--1118.
\bibitem{liu2017gca}
J.~Liu, G.~Wang, L.~Y.~Duan, K.~Abdiyeva, and A.~C.~Kot,
``Skeleton-Based Human Action Recognition with Global Context-Aware Attention LSTM Networks,''
\emph{IEEE Transactions on Image Processing}, vol.~27, no.~4, pp.~1586--1599, 2017.
\bibitem{liu2017stlstm}
J.~Liu, A.~Shahroudy, D.~Xu, A.~C.~Kot, and G.~Wang,
``Skeleton-Based Action Recognition Using Spatio-Temporal LSTM Network with Trust Gates,''
\emph{IEEE Transactions on Pattern Analysis and Machine Intelligence}, vol.~40, no.~12, pp.~3007--3021, 2017.

\bibitem{li2017skelcnn}
C.~Li, Q.~Zhong, D.~Xie, and S.~Pu,
``Skeleton-Based Action Recognition with Convolutional Neural Networks,''
in \emph{2017 IEEE International Conference on Multimedia \& Expo Workshops (ICMEW)}, 2017, pp.~597--600. IEEE.
\bibitem{xin2023transformer}
W.~Xin, R.~Liu, Y.~Liu, Y.~Chen, W.~Yu, and Q.~Miao,
``Transformer for Skeleton-Based Action Recognition: A Review of Recent Advances,''
\emph{Neurocomputing}, vol.~537, pp.~164--186, 2023.

\bibitem{Yan2018Spatial}
S. Yan, Y. Xiong, and D. Lin, ``Spatial temporal graph convolutional networks for skeleton-based action recognition,'' in \emph{Proceedings of the AAAI Conference on Artificial Intelligence}, vol. 32, no. 1, 2018.

\bibitem{Shi2019Two}
L. Shi, Y. Zhang, J. Cheng, and H. Lu, ``Two-stream adaptive graph convolutional networks for skeleton-based action recognition,'' in \emph{Proceedings of the IEEE/CVF Conference on Computer Vision and Pattern Recognition (CVPR)}, 2019, pp. 12026-12035.

\bibitem{Si2019Attention}
C. Si, W. Chen, W. Wang, L. Wang, and T. Tan, ``An attention enhanced graph convolutional LSTM network for skeleton-based action recognition,'' in \emph{Proceedings of the IEEE/CVF Conference on Computer Vision and Pattern Recognition (CVPR)}, 2019, pp. 1227-1236.

\bibitem{Cheng2020Skeleton}
K. Cheng, Y. Zhang, C. Cao, L. Shi, J. Cheng, and H. Lu, ``Skeleton-Based Action Recognition With Shift Graph Convolutional Network,'' in \emph{Proceedings of the IEEE/CVF Conference on Computer Vision and Pattern Recognition (CVPR)}, 2020, pp. 183-192.

\bibitem{Liu2020Disentangling}
Z. Liu, H. Zhang, Z. Chen, Z. Wang, and W. Ouyang, ``Disentangling and unifying graph convolutions for skeleton-based action recognition,'' in \emph{Proceedings of the IEEE/CVF Conference on Computer Vision and Pattern Recognition (CVPR)}, 2020, pp. 143-152.

\bibitem{Chen2021Channel}
Y. Chen, Y. Zhang, Z. Liu, X. Zhang, and T. Han, ``Channel-wise topology refinement graph convolution for skeleton-based action recognition,'' in \emph{Proceedings of the IEEE/CVF International Conference on Computer Vision (ICCV)}, 2021, pp. 1341-1351.

\bibitem{Li2021InfoGCN}
M. Li, S. Chen, X. Chen, Y. Zhang, and Y. Liu, ``InfoGCN: Representation learning for human skeleton-based action recognition,'' in \emph{Proceedings of the IEEE/CVF International Conference on Computer Vision (ICCV)}, 2021, pp. 1331-1340.

\bibitem{Plizzari2021Spatial}
C. Plizzari, M. Cannici, and M. Matteucci, ``Spatial temporal transformer network for skeleton-based action recognition,'' in \emph{International Conference on Pattern Recognition (ICPR)}, 2021, pp. 1011-1018.

\bibitem{Lee2023Hierarchically}
J. Lee, M. Lee, D. Lee, and S. Lee, ``Hierarchically decomposed graph convolutional networks for skeleton-based action recognition,'' in \emph{Proceedings of the IEEE/CVF International Conference on Computer Vision (ICCV)}, 2023, pp. 10444-10453.

\bibitem{maass1997spiking}
W.~Maass,
``Networks of Spiking Neurons: The Third Generation of Neural Network Models,''
\emph{Neural Networks}, vol.~10, no.~9, pp.~1659--1671, 1997.

\bibitem{roy2019spike}
K.~Roy, A.~Jaiswal, and P.~Panda,
``Towards Spike-Based Machine Intelligence with Neuromorphic Computing,''
\emph{Nature}, vol.~575, no.~7784, pp.~607--617, 2019.

\bibitem{schuman2022neuromorphic}
C.~D.~Schuman, S.~R.~Kulkarni, M.~Parsa, \emph{et~al.},
``Opportunities for Neuromorphic Computing Algorithms and Applications,''
\emph{Nature Computational Science}, vol.~2, no.~1, pp.~10--19, 2022.
\bibitem{myung2024degn}
W.~Myung, N.~Su, J.~H.~Xue, and G.~Wang,
``DEGCN: Deformable graph convolutional networks for skeleton-based action recognition,''
\emph{IEEE Trans. Image Process.}, vol.~33, pp.~2477--2490, 2024.

\bibitem{Zhou2024BlockGCN}
Y. Zhou, X. Yan, Z.-Q. Cheng, and Y. Chen, ``BlockGCN: Redefine topology awareness for skeleton-based action recognition,'' in \emph{Proceedings of the IEEE/CVF Conference on Computer Vision and Pattern Recognition (CVPR)}, 2024, pp. 1061-1070.

\bibitem{Wang2025Hulk}
Y. Wang \emph{et al.}, ``Hulk: A universal knowledge translator for human-centric tasks,'' \emph{IEEE Transactions on Pattern Analysis and Machine Intelligence (TPAMI)}, vol. 47, no. 7, p. 10930828, 2025.
\bibitem{Gerstner2014Neuronal}
W. Gerstner, W. M. Kistler, R. Naud, and L. Paninski, \emph{Neuronal Dynamics: From Single Neurons to Networks and Models of Cognition}. Cambridge, U.K.: Cambridge Univ. Press, 2014.
\bibitem{singhal2025snnreview}
T.~Singhal, I.~Rani, D.~Singh, B.~Kumar, V.~Bhatia, and S.~Gupta,
``A Systematic Review of Spiking Neural Networks and Their Applications,''
in \emph{Revolutionizing AI with Brain-Inspired Technology: Neuromorphic Computing}, 2025, pp.~43--60.

\bibitem{Zhou2022Spikformer}
Z. Zhou, Y. Zhou, and R. Li, ``Spikformer: When spiking neural network meets transformer," in \emph{Proc. Eur. Conf. Comput. Vis. (ECCV)}, 2022, pp. 1--18.

\bibitem{Yao2023SpikeTransformer}
M. Yao, J. Hu, Z. Zhou, L. Yuan, Y. Tian, B. Xu, and G. Li, ``Spike-driven transformer," in \emph{Advances in Neural Information Processing Systems (NeurIPS)}, 2023.

\bibitem{Yao2024SpikeV2}
M. Yao, J. Hu, T. Hu, Y. Xu, Z. Zhou, Y. Tian, B. Xu, and G. Li, ``Spike-driven transformer v2: Meta spiking neural network architecture inspiring the design of next-generation neuromorphic chips," \emph{arXiv preprint arXiv:2404.03663}, 2024.

\bibitem{Zou2025SNNViT}
Y. Zou, T. Zhang, and Z. Li, ``Spiking vision transformer with saccadic attention," in \emph{Proc. Int. Conf. Learn. Represent. (ICLR)}, 2025.

\bibitem{Lee2025STAtten}
D. Lee, M. Lee, and S. Lee, ``Spiking transformer with spatial-temporal attention," in \emph{Proc. IEEE/CVF Conf. Comput. Vis. Pattern Recognit. (CVPR)}, 2025.

\bibitem{Zhou2024QKFormer}
C. Zhou, H. Zhang, Z. Zhou, and L. Yu, ``QKFormer: Hierarchical spiking transformer using Q-K attention," in \emph{Advances in Neural Information Processing Systems (NeurIPS)}, 2024.

\bibitem{Fang2024SWformer}
Y. Fang, T. Zhao, and R. Wang, ``Spiking wavelet transformer," in \emph{Proc. Eur. Conf. Comput. Vis. (ECCV)}, 2024.

\bibitem{Zhu2022SpikingGCN}
Z. Zhu, J. Peng, J. Li, L. Chen, Q. Yu, and S. Luo, ``Spiking graph convolutional networks," in \emph{Proc. Int. Joint Conf. Artif. Intell. (IJCAI)}, 2022, pp. 2434--2440.

\bibitem{Yin2024DySIGN}
N. Yin, M. Wang, Z. Chen, G. De Masi, and B. Gu, ``Dynamic spiking graph neural networks," in \emph{Proc. AAAI Conf. Artif. Intell.}, 2024.

\bibitem{Zheng2025MKSGN}
N. Zheng, H. Xia, Z. Liang, and Y. Du, ``MK-SGN: A spiking graph convolutional network with multimodal fusion and knowledge distillation for skeleton-based action recognition," \emph{Neurocomputing}, vol. 568, p. 127048, 2025.

\bibitem{AlFahoum2014EEG}
A. S. Al-Fahoum and A. A. Al-Fraihat, ``Methods of EEG signal features extraction using linear analysis in frequency and time-frequency domains," \emph{ISRN Neurosci.}, vol. 2014, p. 730218, 2014.

\bibitem{Morales2022TimeFrequency}
S. Morales and J. A. Bowers, ``Time-frequency analysis methods and their application in developmental EEG data," \emph{Dev. Cogn. Neurosci.}, vol. 54, p. 101067, 2022.

\bibitem{shahroudy2016ntu}
A.~Shahroudy, J.~Liu, T.-T.~Ng, \emph{et~al.}, ``NTU RGB{+}D: A Large-Scale Dataset for 3D Human Activity Analysis,'' in \emph{Proceedings of the IEEE Conference on Computer Vision and Pattern Recognition (CVPR)}, 2016, pp.~1010--1019.

\bibitem{liu2019ntu}
J.~Liu, A.~Shahroudy, M.~Perez, \emph{et~al.}, ``NTU RGB{+}D 120: A Large-Scale Benchmark for 3D Human Activity Understanding,'' \emph{IEEE Transactions on Pattern Analysis and Machine Intelligence}, vol.~42, no.~10, pp.~2684--2701, 2019.

\bibitem{wang2014cross}
J.~Wang, X.~Nie, Y.~Xia, \emph{et~al.}, ``Cross-View Action Modeling, Learning and Recognition,'' in \emph{Proceedings of the IEEE Conference on Computer Vision and Pattern Recognition (CVPR)}, 2014, pp.~2649--2656.
\bibitem{feng2022gcnreview}
L.~Feng, Y.~Zhao, W.~Zhao, and J.~Tang,
``A Comparative Review of Graph Convolutional Networks for Human Skeleton-Based Action Recognition,''
\emph{Artificial Intelligence Review}, vol.~55, no.~5, pp.~4275--4305, 2022.

\bibitem{liu2016stlstm}
J.~Liu, A.~Shahroudy, D.~Xu, \emph{et~al.}, ``Spatio-Temporal LSTM with Trust Gates for 3D Human Action Recognition,'' in \emph{European Conference on Computer Vision (ECCV)}. Cham: Springer International Publishing, 2016, pp.~816--833.
\bibitem{du2015hierarchical}
Y.~Du, W.~Wang, and L.~Wang, ``Hierarchical Recurrent Neural Network for Skeleton Based Action Recognition,'' in \emph{Proceedings of the IEEE Conference on Computer Vision and Pattern Recognition (CVPR)}, 2015.
\bibitem{lee2025spiketrans}
D.~Lee, Y.~Li, Y.~Kim, S.~Xiao, and P.~Panda,
``Spiking Transformer with Spatial-Temporal Attention,''
in \emph{Proceedings of the IEEE/CVF Conference on Computer Vision and Pattern Recognition (CVPR)}, 2025, pp.~13948--13958.

\bibitem{chi2022infogcn}
H.~G.~Chi, M.~H.~Ha, S.~Chi, S.~W.~Lee, Q.~Huang, and K.~Ramani,
``InfoGCN: Representation Learning for Human Skeleton-Based Action Recognition,''
in \emph{Proceedings of the IEEE/CVF Conference on Computer Vision and Pattern Recognition (CVPR)}, 2022, pp.~20186--20196.

\bibitem{zhou2023ldr}
H.~Zhou, Q.~Liu, and Y.~Wang,
``Learning Discriminative Representations for Skeleton-Based Action Recognition,''
in \emph{Proceedings of the IEEE/CVF Conference on Computer Vision and Pattern Recognition (CVPR)}, 2023, pp.~10608--10617.
\bibitem{2025signalsgn}
``Signal-SGN: A Spiking Graph Convolutional Network for Skeleton Action Recognition via Learning Temporal-Frequency Dynamics,''
\emph{Under review for an anonymous conference submission.}
\bibitem{merolla2014million}
P.~A.~Merolla, J.~V.~Arthur, R.~Alvarez-Icaza, A.~S.~Cassidy, J.~Sawada, F.~Akopyan, \emph{et~al.}, ``A Million Spiking-Neuron Integrated Circuit with a Scalable Communication Network and Interface,'' \emph{Science}, vol.~345, no.~6197, pp.~668--673, 2014.

\end{thebibliography}

\begin{thebibliography}{1}
\bibliographystyle{IEEEtran}
\bibitem{kundu2021spikethrift}
S.~Kundu, G.~Datta, M.~Pedram, and P.~A.~Beerel,
``Spike-thrift: Towards energy-efficient deep spiking neural networks by limiting spiking activity via attention-guided compression,''
in \emph{Proc. IEEE/CVF Winter Conf. Appl. Comput. Vis. (WACV)}, 
2021, pp.~3953--3962.
\bibitem{hu2021advancing}
Y.~Hu, Y.~Wu, L.~Deng, and G.~Li,
``Advancing residual learning towards powerful deep spiking neural networks,''
\emph{arXiv preprint arXiv:2112.08954}, 2021.
\bibitem{kundu2021hiresnn}
S.~Kundu, M.~Pedram, and P.~A.~Beerel,
``HIRE-SNN: Harnessing the inherent robustness of energy-efficient deep spiking neural networks by training with crafted input noise,''
in \emph{Proc. IEEE/CVF Int. Conf. Comput. Vis. (ICCV)}, 
2021, pp.~5209--5218.
\bibitem{kim2021optimizing}
Y.~Kim and P.~Panda,
``Optimizing deeper spiking neural networks for dynamic vision sensing,''
\emph{Neural Networks}, vol.~144, pp.~686--698, 2021.



\end{thebibliography}
\end{document}